\documentclass[journal]{IEEEtran}
\usepackage{amssymb}
\usepackage{multirow}
\usepackage{amsthm}
\usepackage{mathrsfs}
\usepackage{setspace}
\usepackage{enumitem}
\usepackage{array}
\usepackage{float}
\usepackage{subfigure}
\usepackage{booktabs}
\usepackage{amsmath}
\usepackage{boxedminipage}
\usepackage{graphicx}
\usepackage{gensymb}
\usepackage{color}

\usepackage{cite}

\ifCLASSINFOpdf
\else
\fi

\begin{document}
%
\title{Explicit3D: Graph Network with Spatial Inference for Single Image 3D Object Detection}
%
%
%

\author{Yanjun~Liu,
        and~Wenming~Yang*,
\thanks{*Corresponding author: Wenming Yang.}
\thanks{Yanjun Liu, Wenming Yang are with the Tsinghua Shenzhen International Graduate School, Tsinghua University, Shenzhen, Guangdong, 518000 China. E-mail: liuyanju21@mails.tsinghua.edu.cn, yang.wenming@sz.tsinghua.edu.cn.}
}

%
%

\markboth{IEEE TRANSACTIONS ON MULTIMEDIA, ~Vol.~0, No.~0, June~2023}%
{Liu \MakeLowercase{\textit{et al.}}: Graph Network with Spatial Inference for Single Image 3D Object Detection}
%



\maketitle

\begin{abstract}
Indoor 3D object detection is an essential task in single image scene understanding, impacting spatial cognition fundamentally in visual reasoning. Existing works on 3D object detection from a single image either pursue this goal through independent predictions of each object or implicitly reason over all possible objects, failing to harness relational geometric information between objects. To address this problem, we propose a dynamic sparse graph pipeline named Explicit3D based on object geometry and semantics features. Taking the efficiency into consideration, we further define a relatedness score and design a novel dynamic pruning algorithm followed by a cluster sampling method for sparse scene graph generation and updating. Furthermore, our Explicit3D introduces homogeneous matrices and defines new relative loss and corner loss to model the spatial difference between target pairs explicitly. Instead of using ground-truth labels as direct supervision, our relative and corner loss are derived from the homogeneous transformation, which renders the model to learn the geometric consistency between objects. The experimental results on the SUN RGB-D dataset demonstrate that our Explicit3D achieves better performance balance than the-state-of-the-art.
\end{abstract}

\begin{IEEEkeywords}
Scene Understanding, 3D Object Detection, Graph Neural Network, Homogeneous Transformation.
\end{IEEEkeywords}

%
\IEEEpeerreviewmaketitle

\section{Introduction}
%
%
%
%
Given an RGB image, the task of indoor scene understanding is to model common indoor objects (usually furniture) and estimate their semantic category. Understanding indoor scenes has shown unique importance in various applications such as digital twin, interior design, robot navigation, and content synthesis in augmented and virtual reality. However, with the absence of depth information, inferring 3D space from 2D images is inherently ill-posed because scenes with distinct geometry structures can be projected onto identical RGB images. Besides, heavily occluded scenes even make things worse by filtering out useful pixel information for estimating position, orientation, and scale. Furthermore, in scenes with significant variations in object poses and scales, it can be challenging work to identify different objects in the real world, even for human brains. Thus, the general performance of indoor scene understanding is still far from satisfactory.

\begin{figure}
	\centering
	\includegraphics[width=\linewidth]{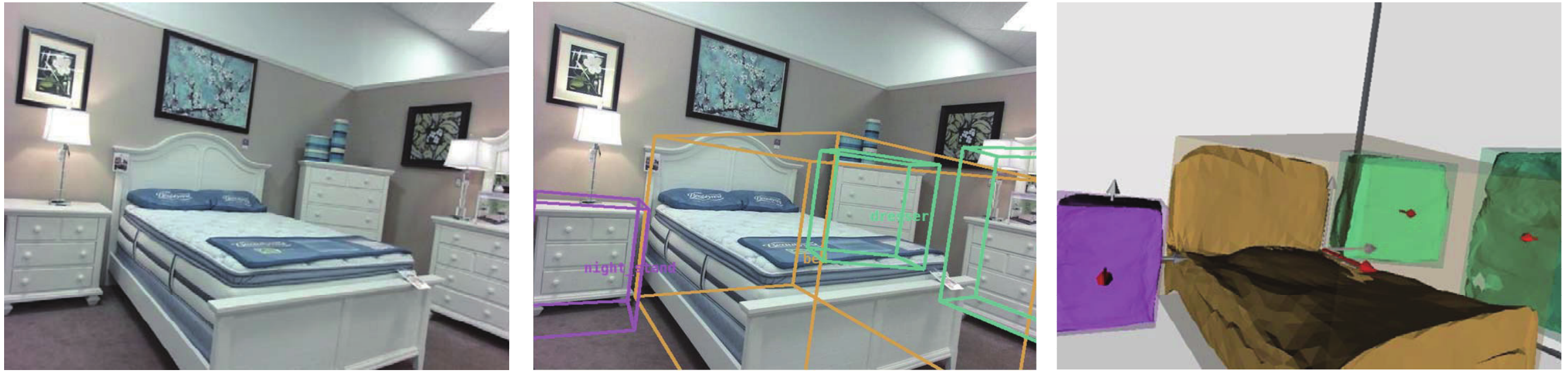}
	\caption{Indoor scene understanding takes a single image (left) to perform 3D object detection (middle) and semantic reconstruction (right).}
    \label{figbedroom}
\end{figure}

Over decades, researchers have been trying to consistently improve scene understanding algorithms by designing more accurate 3D object detection systems \cite{lian2022exploring, liu2022learning, zhang2022dimension} and introducing object-centric shape reconstruction \cite{genova2020local, nie2020total3dunderstanding, mildenhall2020nerf}. The former task mainly tackles the problem of locating the center of relevant objects and predicting their spatial occupancy and orientation. On the other hand, the latter task is to reconstruct and render the objects' shape and surface details. This paper focuses on indoor 3D detection, which is essential for further scene understanding tasks such as shape reconstruction, semantic reconstruction, and comprehensive scene rendering and optimization (as shown in Fig. \ref{figbedroom}).

A well-used approach for indoor object detection is to model relevant objects in a bounding cuboid manner \cite{huang2018cooperative,choi2013understanding} and the regression parameters will be the coordinates of the cuboid center, cuboid scale, and also its rotation angles. The problem will be straightforward for RGB-D images or point clouds under the projection of a pin-hole camera model. However, for single-image scene object detection tasks, vast and intuitive depth information cannot be sensed from real-world environments. Thus, recent works are focusing on utilizing geometric \cite{lian2022exploring}, pixel-level \cite{nie2020total3dunderstanding, huang2018holistic} and semantic \cite{zhang2021holistic} attributes in RGB images to better assist the detection task.

Although existing object-centric 3D detection algorithms significantly matured over the last decade \cite{chen2019holistic++,avetisyan2020scenecad,shi2020point}, most of them represent indoor scenes as collection of independent objects. The unique characteristic of the indoor scene detection task is that there are usually several relevant objects appearing simultaneously in a single image, and the frequency and geometry of the co-occurrence with each other can be probabilistically modeled. The insight is inspired by the theory pointed out by \cite{kulkarni20193d} that human brains rely on visual priors at object-level, scene-level, and relationship-level when understanding indoor scenes. Take Fig. \ref{figbedroom} as an example, if there is a bed in the scene (object-level), we can make bold assumptions that there may also be nightstands next to the bed (relationship-level), and this scene is most likely to be a bedroom (scene-level). Since we have identified the problem as bedroom scene understanding, there may also be wardrobes, dressing table \emph{etc.} in the image (relationship-level). Thus, it is possible to harness these probabilistic relationships between different objects to boost the performance of our detection system.

So how to model co-occurring frequencies for 3D object detection becomes a critical issue. The natural idea is to fuse all pairwise information for object prediction \cite{nie2020total3dunderstanding, kulkarni20193d}. Since dense pairwise connections can also be interpreted as graphs, the methods above actually reason over fully connected graphs where each object-pair relationship is a potential edge. While a fully connected graph contains all pairwise relationships, it scales quadratically with the number of objects, rendering inference over large and complex scenes quickly becomes impractical. Couples of works \cite{zhou2019scenegraphnet, yang2018graph} have tried to adopt sparse scene graph models to understand scenes. The state-of-the-art work \cite{zhang2021holistic} also employs GCN-based scene graphs to aggregate object-level and contextual information in an implicit way and achieves dramatic improvements compared to dense graph-based methods, which indicates the potential of harnessing sparse scene graphs. One of the possible reasons is that by designing an appropriate attention mechanism or aggregation scheme, sparse graphs focus on more useful relational information and perform more efficient inference than dense graphs.

On the other hand, although \cite{zhang2021holistic} addresses the problem of sparse graph learning, it still needs to model shape features and contextual information in an implicit representation. In other words, implicit manners allow neural networks to learn latent information such as spatial translation and distance between different objects without explicitly considering the probabilistic geometric arrangement mentioned above. 

Thus, taking advantage of the interpretability of explicit spatial relationships and the potential of sparse graphs, we propose a novel sparse graph neural network-based method where probabilistic relationships are modeled as explicit geometric transformation matrices. In particular, we propose a dynamic graph pruning module based on cluster sampling method, which calculates the pairwise relatedness score through the interaction between geometry and semantics, and an edge message update scheme to predict the explicit spatial information between target pairs. With the above settings, we demonstrate the performance-cost balance of our graph network-based pipeline on the SUN RGB-D dataset and also extend our algorithm to holistic scene modeling tasks. In summary, our contribution is four-fold:
\begin{itemize}[leftmargin=10pt]
	\item We propose a sparse graph-based pipeline for single image 3D indoor scene detection. The node embedding and edge message in graph network are integrated by geometric constraints an for joint supervised learning.
	\item Based on our novel relatedness score, a dynamic pruning method is purposed for sparse scene graph generation. We design the cluster sampling algorithm to reduce computational cost while guaranteeing the prediction accuracy.  
	\item The homogeneous matrix is used to represent the relative spatial information in an explicit manner. We are the first work to model the relationship between objects in a geometric manner for 3D detection problem.
	\item Based on the principle of homogeneous transformation, our method is also able to learn relative spatial parameters through novel relative loss and corner loss.   
\end{itemize}


\section{Related Works}
\subsection{3D Object Detection} 
The task of 3D object detection is similar to 2D detection but lifts the 2D boxes to 3D space. Over the years, researchers have explored several commonly used data formats for 3D detection, including depth image, voxel, multi-view images, and point cloud data. Existing works can be divided into three main categories according to different data inputs: RGB images, point cloud, and hybrid data fusion. Many works \cite{Patil2022CVPR, lian2022exploring} have been trying to recover the 3D space from merely 2D images, and their focus is to perform detection tasks with the help of estimated depth images. Detection on point cloud \cite{qi2017pointnet, shi2020point, shi2020pv, shi2019pointrcnn, tmm22_detection} extends pixel in 2D space to point cloud in 3D space, which processes point cloud directly without projection or segmentation. Data fusion methods \cite{chen2017multi, qi2018frustum} integrate RGB images and point cloud from LiDAR for 3D object detection, taking advantage of mature 2D object detectors to narrow down the proposal search space. They extract the 3D bounding boxes by aligning 2D ones from image detectors to 3D space. In this paper, the goal of our scene understanding task is to directly predict the bounding boxes in 3D space from a single RGB image instead of explicitly estimating the miss depth information.

\subsection{Semantic Reconstruction} 
Indoor semantic reconstruction aims to reconstruct the whole scene holistically with shape details. Concretely, common reconstruction pipelines are often concerned with two topics: scene understanding and shape representation. 

Scene understanding task is mainly about understanding and geometrically interpreting indoor scenes without considering shape details, which includes room layout estimation, object classification, localization \emph{etc.} Researchers in this field have consistently proposed different approaches to predict the indoor scene geometry \cite{tmm20_layout, tmm21_layout}, and object locations from a single image \cite{chen2019holistic++, huang2018holistic, choi2013understanding}.

The shape representation task concentrates more on generating 3D shapes corresponding to the input images, which can be further categorized into three categories: scene-level reconstruction, object-wise reconstruction, and model-retrieval methods. Scene-level methods recover the 3D texture of the entire scene in an end-to-end manner without explicitly modeling each object in the scene. Most scene-level reconstruction methods adopt depth or voxel information \cite{shin20193d, wu2020pq, xu2019disn} to represent the whole scene. However, these algorithms often suffer from computational inefficiency, and their reconstruction resolution is also limited. Model retrieval methods \cite{izadinia2017im2cad} mainly utilize template matching methods to retrieve CAD models which share the most similarity with the input images from the given model repository. The advantage of this sort of method is their improved shape quality over those scene-level counterparts. However, in model retrieval approaches, only CAD models in a given repository can be matched with input images, which severely limits its practical application. Object-wise reconstruction methods, on the other hand, can be more accurate and flexible since it relies entirely on pixel information of the target object and recovers the target surface in an object-centric coordinate system. One commonly used shape representation is called triangular mesh and in recent years,  mesh reconstruction methods \cite{wang2018pixel2mesh, gkioxari2019mesh, pan2019deep, Park2019CVPR, Li2022CVPR} have exhibited advantages in accuracy and efficiency. 

In our work, we mainly concentrate on indoor scene understanding tasks, especially 3D object detection based on a single image. Object-wise semantic reconstruction based on the triangular mesh will also be employed as an application scenario to evaluate the performance of our 3D detection algorithm.

\begin{figure*}[htbp!]
	\centering
	\includegraphics[width=\linewidth]{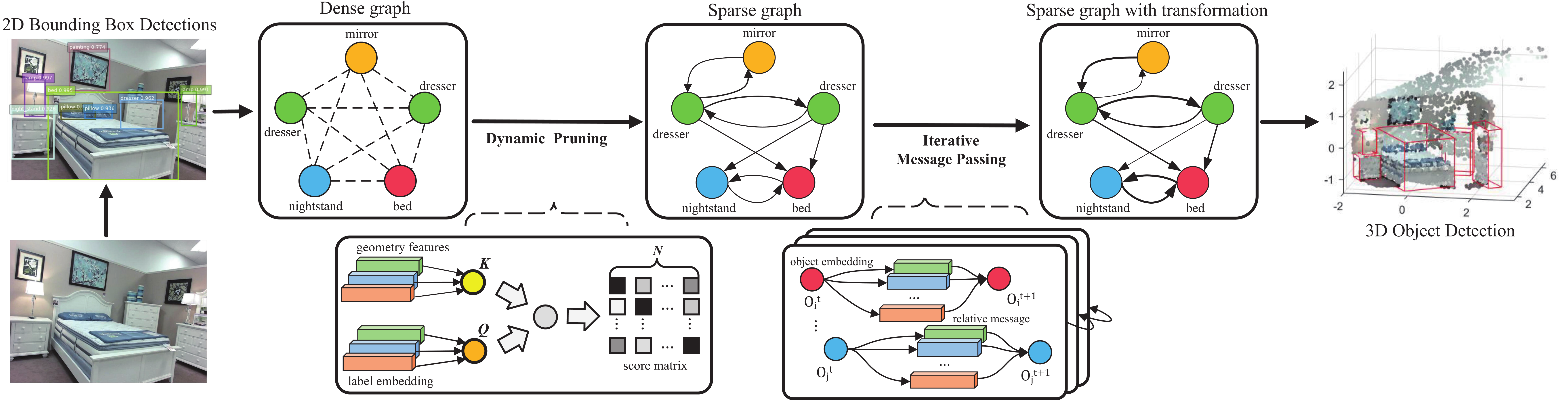}
	\caption{Overview of the pipeline of our graph neural network.}
	\label{figoverview}
\end{figure*}

\subsection{Scene Graphs Networks}
Scene graph was initially used in computer graphics as a flexible representation to arrange spatial occupancy of a graphical scene until Johnson \emph{et al.} \cite{johnson2015image} re-proposed it to model scenes as graphs. The problem of generating scene graphs from images extends object detection and instance segmentation \cite{chen2019holistic++, liu2016ssd, redmon2016you, he2017mask} to relationship detection and transformations of object pairs. A typical scene graph comprises objects as nodes and pairwise edges, where nodes represent instance information, and edges can represent low-level spatial information and high-level structural relationships such as affordance, supporting, surrounding, \emph{etc.}, between object pairs. The structural graph itself provides some of the spatial prior for supervised learning. Thus it is inherently an ideal implementation of graph neural networks. Some works have attempted to use scene graph networks to address 3D scene understanding problems such as indoor 3D object detection \cite{zhou2019scenegraphnet} , scene semantic segmentation \cite{wang2019graph}, position-based object recognition \cite{wald2020learning}, \emph{etc.} But these methods either extract 3D scene graphs directly from point clouds based on solid human spatial priors or construct a 2D scene graph by scene graph generation algorithms before inferring 3D information, which involves too much painstaking manual work to parse scenes. Besides, there have not been datasets based on graph structure designed for single-image 3D object detection. Thus, instead of specifying the high-level relationship categories (\emph{i.e.}, wearing on, has, holding \emph{etc.}), we focus on low-level spatial intuition, transformations including translation and rotation, to be specific. Thus, no scene graph generation problem or graph pre-processing needs to be concerned in this paper.

Another topic that researchers are concerned about is the design of graph neural network structure. Since it is commonly recognized that reasoning over a fully connected graph with the quadratic number of connections is straightforward but computational costly, many methods recently have turned to model scenes as graph neural networks with sparse dependencies amongst objects \cite{battaglia2016interaction, scarselli2008graph, zhang2020deep, gilmer2017neural, kipf2016semi}. This paper focuses on graph neural networks where each node is sparsely connected with multiple edges across objects. Our graph network performs message passing along edges and updates node-wise embedding through information aggregation operations. In other words, the 3D object detection process here is regarded as an iterative optimization of the graph structure.

\section{Method}
The overview of our method is illustrated in Fig. \ref{figoverview}. The entire detection pipeline consists of graph pruning, edge message update and the prediction of 3D pose. Given an image, our model first performs 2D object detection as in \cite{zhang2021holistic, nie2020total3dunderstanding}. All detected bounding boxes constitute a dense scene graph with all nodes connecting with each other through bilateral edges. Then our algorithm prunes the connections between objects through the relatedness score matrix, which yields a sparse graph afterwards. Edge message update scheme over sparse graph is then applied to aggregate multi-lateral messages between neighbouring nodes for each object. Finally, we obtain the 3D pose of objects by decoding the node embedding and transformation prediction from the sparse graph model. We firstly introduce the 3D object detection formulation in both world system and camera system in Sec. \ref{coord}. We then present our dynamic graph pruning module in Sec. \ref{prune}. We describe our graph neural network model architecture and edge message update scheme in Sec. \ref{graph}. In Sec. \ref{decoder-loss}, we discuss the independent decoder and relative decoder and how transformation loss is combined with the individual prediction.

\subsection{Object Detection Formulation and Coordinate System \label{coord}}
\begin{figure}[!htbp]
	\centering
	\includegraphics[scale=1]{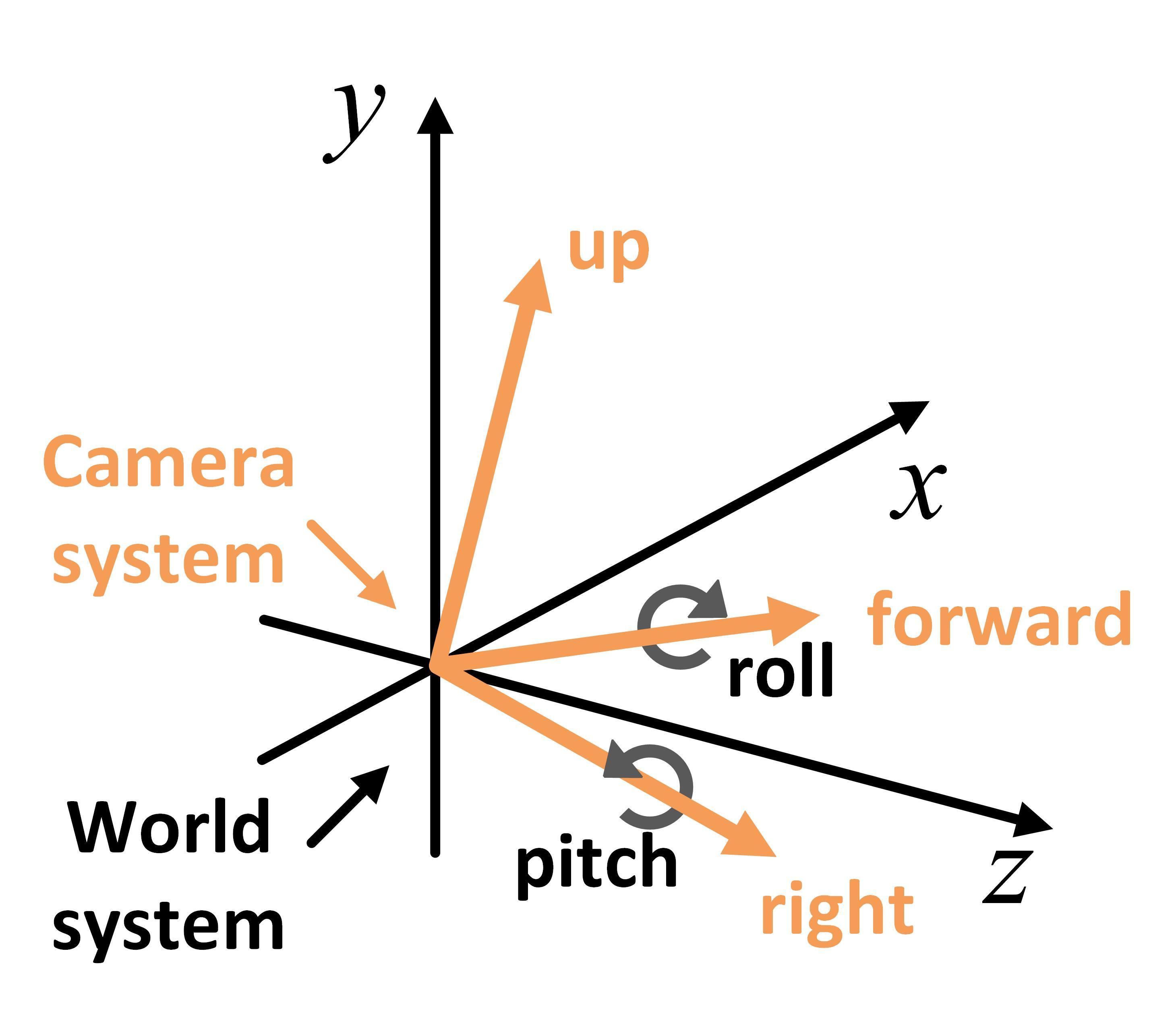}
	\caption{Camera pose and world systems}
	\label{figpose}
\end{figure}

To better represent the indoor scene in the mathematical sense, we firstly parameterize the scene in a box-in-box manner as \cite{nie2020total3dunderstanding}. As shown in Fig. \ref{figpose}, the world and camera coordinate systems share the same origin, where $Y$-axis in world coordinate system is defined as vertical direction perpendicular to the floor. To transform objects in camera to world system, we perform rotation with respect to static axis, i.e., rotate the world system around its $Y$-axis to align the $x$-axis toward the forward direction of the camera, such that the camera’s yaw angle can be removed. Then the camera pose relative to the world system can be expressed by the angles of pitch $\beta$ and roll $\gamma$:

We assume that all the indoor objects stand on the floor supported by the room layout or other objects in the scene. Thus, the pitch-roll-yaw rotation of an arbitrary object in the free space degrades to the orientation along the perpendicular axis.

\begin{figure}[hb]
\centering
\includegraphics[width=0.9\linewidth]{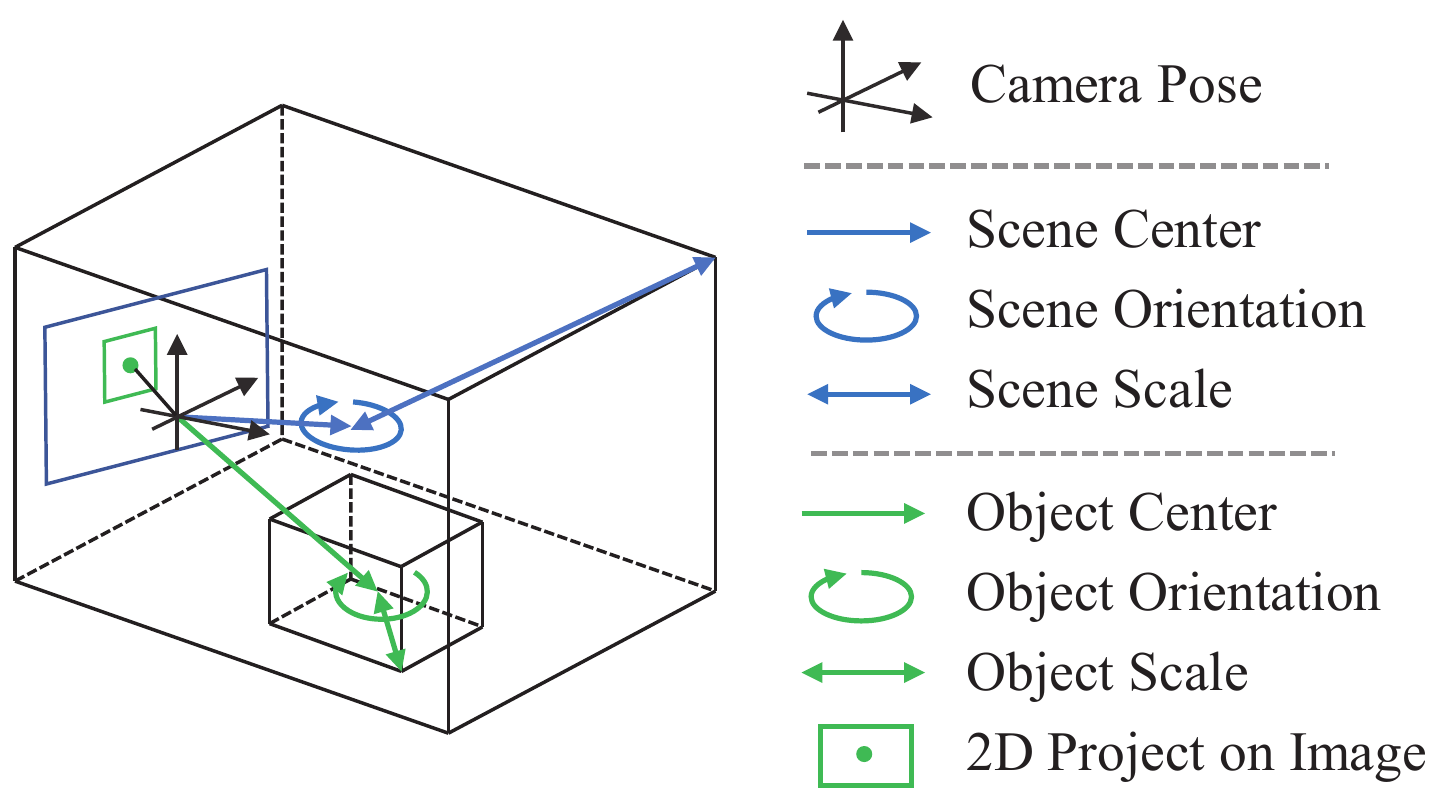}
\caption{Scene  parameterization in box-in-box manner.}
\label{figbox}
\end{figure}

\begin{equation}
R(\beta, \gamma) =
\begin{bmatrix}
& cos(\beta) &-cos(\gamma)sin(\beta) & sin(\beta)sin(\gamma)   \\ 
& sin(\beta) & cos(\beta)cos(\gamma)  &  -cos(\beta)sin(\gamma) \\
& 0 & sin(\gamma) & cos(\gamma)   \\
\end{bmatrix}
\end{equation}

In the world system, a 3D bounding box is represented by a 3D centroid $C \in \mathbb{R}^3$, spatial scale $S \in \mathbb{R}^3$ and orientation angle $\Theta \in [-\pi, \pi)$. For indoor objects in the image, 3D boxes should be constructed from 2D boxes' centers to maintain the consistency between 2D space and 3D space. Thus, $C$ can be represented by its 2D projection $c \in \mathbb{R}^2$ on the image plane with its distance $d \in \mathbb{R}$ to the camera center. Given the camera intrinsic matrix $K \in \mathbb{R}^3$, the 3D centroid can be formulated by:
\begin{equation}
\label{eqproj}
	C = R^{-1}(\beta, \gamma) \cdot d \cdot \frac{K^{-1}[c, 1]^T}{\left \| K^{-1}[c, 1]^T \right \|}
\end{equation}
where $R(\beta, \gamma)$ is rotation matrix of the camera pose \emph{w.r.t} the world coordinate system, named the camera extrinsic matrix and $(\beta, \gamma)$ denotes the pitch and roll angles. The 2D projection center $c$ can be further decoupled into $c^b+\delta$, where $c^b$ is the 2D bounding box center and $\delta \in \mathbb{R}^2$ is the offset to be learned. It should be noted that while in the 3D object detection task, we output the scale, translation and orientation of an object in camera frame. In semantic reconstruction task, we also need to predict camera pose because it is vital for maintaining the consistency between the layout and indoor objects. Specifically, from the 2D detection to its 3D bounding box corners, our network needs to learn such a function that $f(I|\delta, d, \theta) \in \mathbb{R}^{3 \times 8}$ for object detection task and $F(I|\delta, d, \beta, \gamma, s, \theta) \in \mathbb{R}^{3 \times 8}$ for semantic reconstruction task.

\begin{figure*}[htbp]
	\centering
	\includegraphics[width=\linewidth]{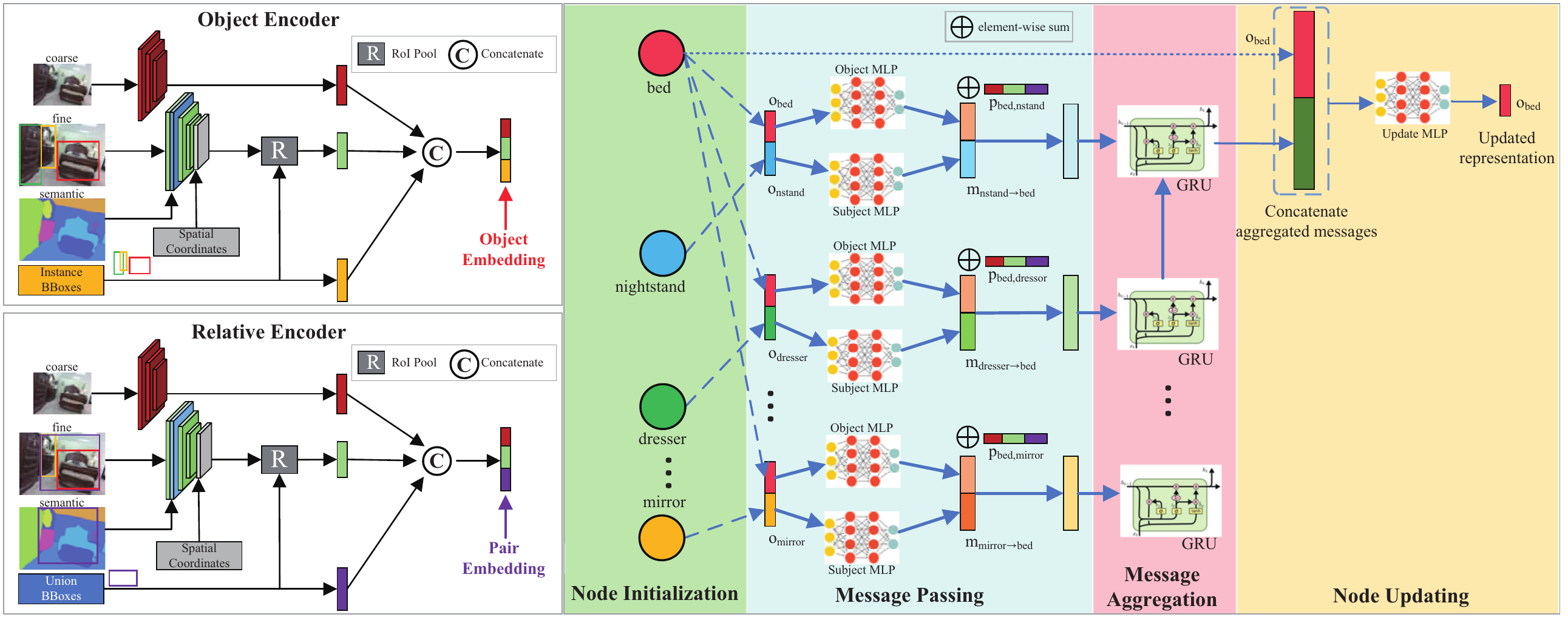}
	\caption{The message passing scheme. Object Encoder (upper left) encodes the per-object embedding, Relative Encoder (bottom left) encodes the union of pairwise bounding box pair as the initial state of graph network. The graph network (right) iteratively compute the object and relative embedding for the prediction through message passing scheme.}
	\label{figmsg}
\end{figure*}

\subsection{Cluster Sampling and Graph Pruning \label{prune}}
From 2D object detection, we obtain the bounding box representation of all the objects in the scene. We initially arranged all the objects in the form of a fully connected scene graph where each object can interact with the other. Although the nodes in a scene graph may have multiple categories of relations between them \cite{zhou2019scenegraphnet}, we do not explicitly model the semantic type of relationship in our graph model and only focus on the spatial aspect. We assume the relationship between a pair of objects is unary and bidirectional.Then the fully connected scene graph can be regarded as a $n \times n$ matrix with all elements 1, where $n$ is the number of objects in the image. However, dense graphs can be inefficient when reasoning over large graphs since, with the growing number of objects, the number of connections grows quadratically \cite{yang2018graph}. Thus, we introduce the relatedness score to control the connection density for graph pruning.

Each 2D bounding box $i$ is associated with a geometry representation $g_i=[x_i, y_i, w_i, h_i]$ and a class label. We score over all directional pairs (leave out identity connection), $n \times (n-1)$ in total, and represent the relatedness between object $i$ and object $j$ as $r^{ij}$. The computation of this score is made up of geometry and semantic weight and contains two steps. 

Firstly, an augmented 4-dimensional geometry representation $g_{ij}^A$ is designed as Eq. \ref{eqpos} to maintain invariance to translation and scale transformations for object pairs. 
\begin{equation}
\label{eqpos}
g^A_{ij} = [log(\frac{|x_i-x_j|}{w_i}), log(\frac{|y_i-y_j|}{h_i}), log(\frac{\omega_i}{\omega_j}), log(\frac{h_i}{h_j})]
\end{equation}
where $x, y$ denotes the center coordinates of the bounding box in pixel space and $w, h$ denotes the width and height of the bounding box. Then this vector is embedded into high-dimensional geometry feature $\varepsilon^{ij}$ using positional encoding \cite{vaswani2017attention, hu2018relation}, which concatenates the sinusoidal and cosinusoidal functions of different wavelengths. Since the geometry weight $\omega_G^{ij}$ in relatedness score only focuses on uni-directional information, it can be defined as the ReLU function of a linear transformation of embedded geometry feature $\varepsilon_{ij}$, where $W_G$ is a learnable transformation matrix.
\begin{equation}
	\omega_G^{ij}=ReLU(W_G \cdot \varepsilon_{ij})
\end{equation}

Secondly, the semantic weight $\omega^{ij}_C$ integrates class label information to infer pairwise relatedness. Concretely,  it is computed as the cosine distance between the word embeddings of label $E_i$ and $E_j$, obtained by looking up the embedding dictionary from BERT \cite{devlin-etal-2019-bert} language model, as in Eq. \ref{eqlabel}.
\begin{equation}
\label{eqlabel}
\omega_C^{ij} = \frac{E_i \cdot E_j}{\left \| E_i\right \| \cdot \left \|E_j\right \|}
\end{equation}

The relatedness score indicates the impact from other nodes, which can be computed as Eq. \ref{eqrelate} and relatedness score matrix is represented as $S=\left\{ r^{ij}\right\}_{i,j=1,2,\cdots n}$. We normalize the relatedness score in $S$ element-wisely to $[0,1]$ via a sigmoid function and sort the scores in descending order. 

\begin{figure}[htbp]
\includegraphics[width=0.8\linewidth]{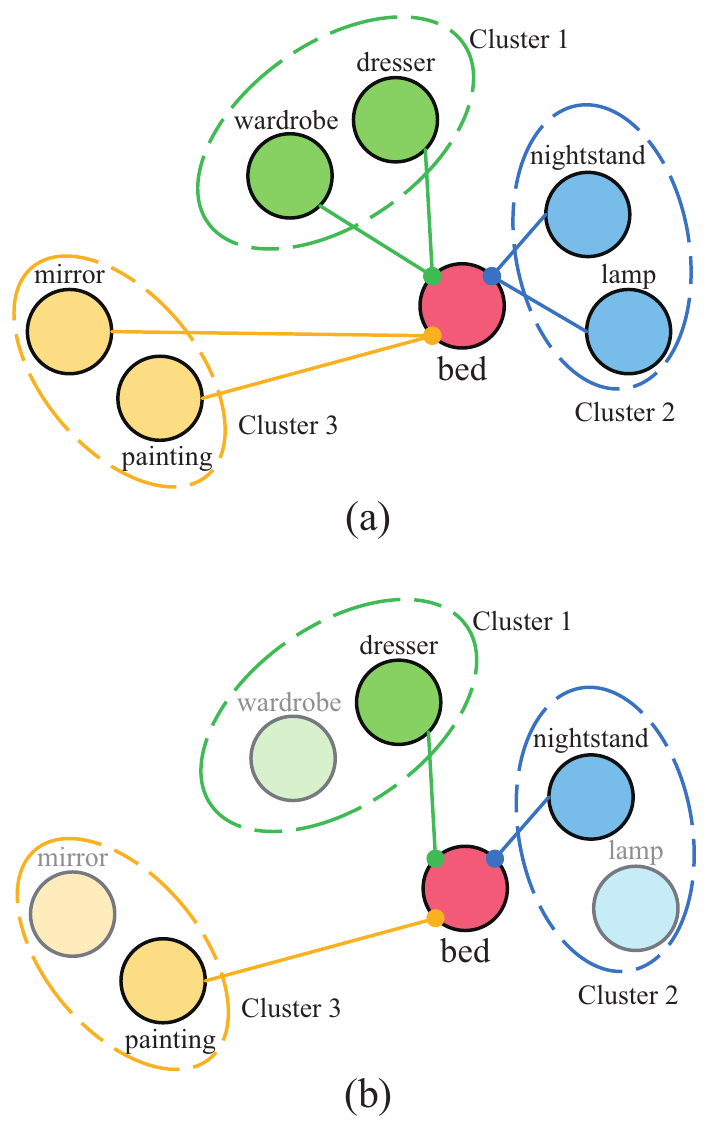}
\caption{Dynamic Graph Pruning: (a) Before Pruning (b) After Pruning. The distances from objects to "bed" are inversely proportional to the relatedness. The yellow, green and blue lines respectively represent farthest, second farthest and nearest distances from clusters to center.}
\label{figrelate}
\end{figure}

Instead of applying an absolute threshold to prune low-relevance connections, we proposed a straight forward pruning method called cluster sampling. This algorithm firstly clusters all relatedness scores into $K$ sets and keep the connections corresponding to the maximum relatedness in each range. As shown in Fig. \ref{figrelate}, in this example scene, when pruning the bed-centered connections, the rest of the objects are divided into three clusters based on their relatedness scores. We pick the connection with the highest relevance for each cluster and eliminate the other connections since objects with similar scores contribute similarly to the final estimate. This way, the graph can capture both short-term and long-term relationships without repeatedly computing the similar object features within the same range.
\begin{equation}
\label{eqrelate}
r^{ij} = \frac{\omega^{ij}_G \cdot exp(\omega^{ij}_C)}{\sum\limits^n\limits_{k=1} \omega^{kj}_G \cdot exp(\omega^{kj}_C)}
\end{equation}

\subsection{Edge Message Passing Scheme\label{graph}}
In the sparse scene graph model constructed in Sec. \ref{prune}, nodes are logically connected by the relatedness score. In this section, an iterative graph update and optimization method is proposed. In our graph network, nodes are inserted as embedding features, and edges are represented as message vectors, as shown in Fig. \ref{figmsg}. During iterations, the nodes emit their state information to other nodes while simultaneously receiving messages from others to understand its context.In this way, the node embeddings can capture self-independent attributes and scene context information. At the same time, the messages on the edges encode the interaction and relationship between the corresponding nodes. Therefore, our graph model can learn to predict both object-wise representations and pairwise relations through message passing and graph iterations. The graph iteration contains initialization, passing, aggregation, and updating steps, which will be elaborated below.

\subsubsection{Node Initialization}
Node embeddings and edge messages have to be initialized so that the iteration process of the scene graph network can be activated. Then the graph can be updated and optimized based on the passing and update scheme discussed later. To start with, we introduce some notation and terminology to simplify the exposition. The node $i$'s embedding is initialized as $o_i$, node $j$'s embedding as $o_j$ and the pairwise embedding between them as $p_{ij}$. The edge message between these two nodes is denoted as $m_{ij}$. 

In our implementation, initial object embeddings $o_i, o_j$ and pair embedding $p_{ij}$ are obtained by object encoder and relative encoder as shown in Fig. \ref{figmsg} respectively. The Faster-RCNN-like object encoder captures coarse-to-fine texture features of objects with awareness of spatial coordinates, while the relative encoder follows a similar architecture but deals with the union of pairwise bounding boxes. Note that the node initialization steps mentioned above only apply to the first iteration.

\subsubsection{Message Updating}
After obtaining the new node embeddings (either after node initialization or update), we need to compute edge messages to perform graph updates and optimization. The edge messages here should be a vector closely related to the difference of pairwise nodes because, in the subsequent prediction task, we need to decode the edge messages for the prediction of relative spatial transformations between objects. Thus, our edge message is derived from both node embeddings and pairwise embeddings. Firstly, the edge message $m_{ij}$ is computed as the concatenation of the object projection $\phi(o_i)$ and subject projection $\psi(o_j)$. We use two asymmetric MLPs with identical architecture for the projection functions $\phi(\cdot)$ and $\psi(\cdot)$, as shown in Fig. \ref{figmsg}. Then we incorporate the pairwise embeddings linear transformed by $W_V$ with the concentration by applying element-wise sum operation, as Eq. \ref{eqmsg}. During graph iteration, pairwise embedding $p_{ij}$ remains a constant vector, but $W_V$ is learnable over time. Thus this dynamic element-wise summation is capable of fusing pairwise information with independent object embeddings.
\begin{equation}
\label{eqmsg}
m_{ij} = SUM(concat(\phi(o_i), \psi(o_j)), W_V \cdot p_{ij})
\end{equation}

\subsubsection{Message Aggregation} 
The message aggregation scheme is designed to fuse context information in the scene and update the entire graph model. The core problem of aggregation is to reduce the dimension of the edge vector while at the same time preserving as much information as possible. In our implementation, the GRU unit is proposed to tackle this problem. Assuming the node $i$ has $n$ edges in total, all edge messages for node $i$ are ordered according to the relatedness scores $r_{ij}, j=1,2,\cdots, n$. Then edge messages are aggregated through GRU units from the last message $m_{in}$ to the first message $m_{i1}$ as Eq. \ref{eqagg}. The typical GRU structure used here takes the last state $h_{ij+1}$ and edge message $m_{ij}$ as input and outputs the current state $h_{ij}$. In this way, the aggregated message can capture the long-term context relationships.
\begin{equation}
\label{eqagg}
\begin{aligned}
& h_{in} = GRU(0, m_{in})  \\
\rightarrow ... \rightarrow & h_{ij} = GRU(h_{ij+1}, m_{ij})  \\
\rightarrow ... \rightarrow & h_{i} = GRU(h_{i2}, m_{i1}) \\
\end{aligned}
\end{equation}

\subsubsection{Node Updating}
After executing message updates and aggregation, the graph network completes one iteration. The node embeddings need to be updated to continuously place excitation on the following graph network iteration. For the node $i$, its embedding is updated as the concatenation of the shortcut node embedding $o_i$ and the output of the last GRU unit $h_i$, as shown in Eq. \ref{equpdate}. It should be noted that edge messages and node embeddings are updated synchronously. Thus for the last iteration, we only select the updated node embeddings and the un-updated edge messages as the input of the subsequent decoders.
\begin{equation}
\label{equpdate}
o_i = MLP(concat(o_i, h_{i}))
\end{equation}

\subsection{Prediction\label{decoder-loss}}
\begin{figure}[htbp]
\includegraphics[width=\linewidth]{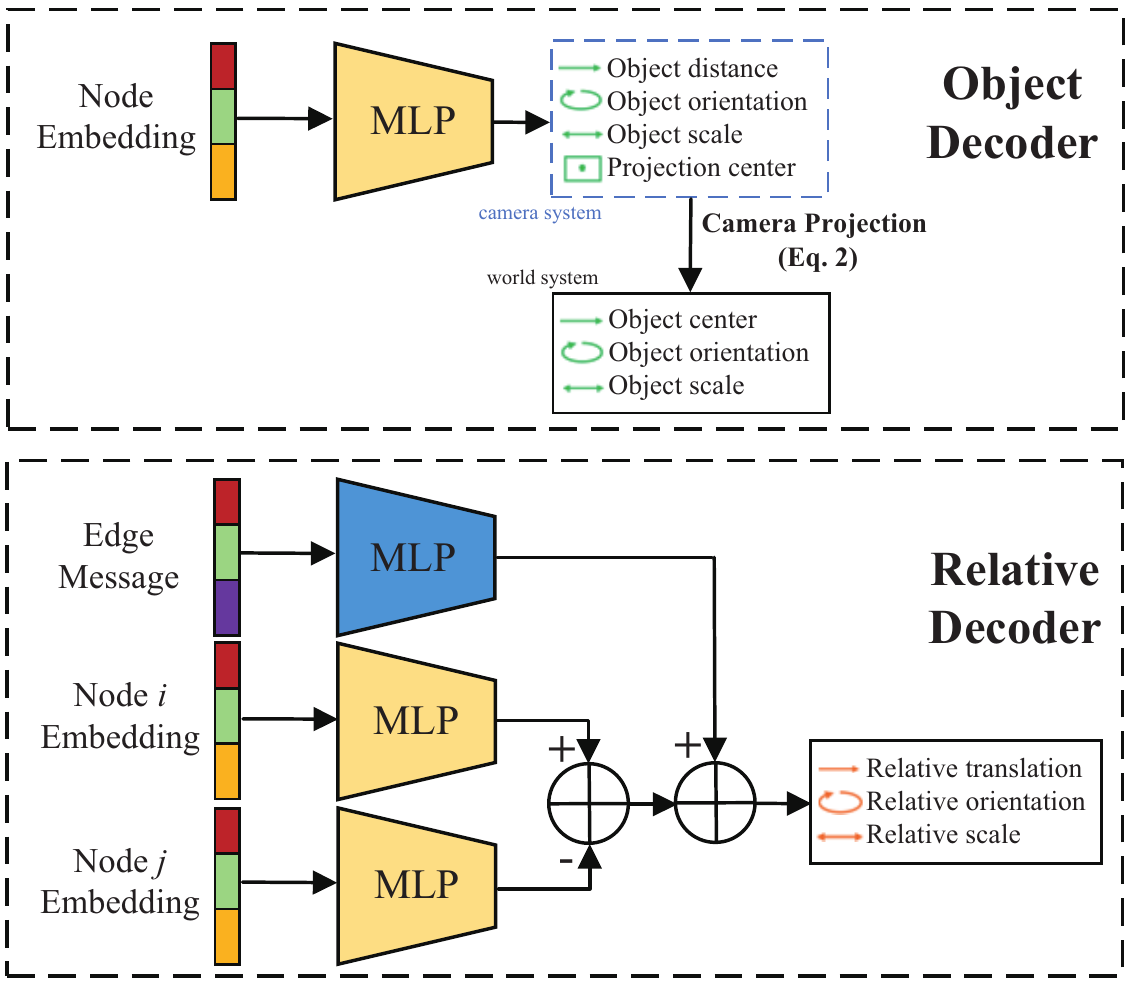}
\caption{Architecture of object decoder and relative decoder.}
\label{figdecoder}
\end{figure}

After the graph iterations discussed in Sec. \ref{graph}, object-wise node embeddings and pairwise edge messages are used for decoding per-object pose and per-pair spatial transformation, respectively. As illustrated in Fig. \ref{figdecoder}, the object decoder uses node embedding to predict depth $d$, scale $s$, orientation $\theta$, and the projection offset $\delta$ in the camera coordinate system  for each object independently. These camera-space parameters can be projected to world-system parameters including bounding box pose $C, \Theta$ and scale $S$ via Eq. \ref{eqproj}. Note that the homogeneous transformation holds distance invariance, so the scale in the camera system and world system remains the same. Orientation in the world system can also be easily figured out by multiplying object frames by the camera rotation matrix, elaborated as Eq. \ref{eqworld_camera}. The camera rotation matrix can either be prior knowledge or estimated using layout estimation network discussed in Sec. \ref{expfigsec}. As the final estimates and loss functions are all defined in the world coordinate system (except for independent loss), we take the world system as the  default reference coordinate below. 

\begin{equation}
\label{eqworld_camera}
\begin{aligned}
& C_{world} = Projection(d_{camera}, c_{image}, R_{camera}) \\
& S_{world} = s_{camera} \\
& R_z(\Theta_{world}) = R_{camera} R_z(\theta_{camera}) \\
\end{aligned}
\end{equation}

The relative decoder takes pairwise edge messages and a pair of node embeddings to form a "filtered" pose signal. It outputs spatial transformation of relative transformation of center $\Delta C_{ij}$, relative log-scale $\Delta S_{ij}$ , and relative orientation $\Delta \Theta_{ij}$. Since all indoor objects are on the floor, they can only rotate along the $Z$-axis. Thus the relative orientation error can be represented by $\Delta \Theta_{ij}=\Theta_i-\Theta_j$. The relative center coordinate $\Delta C_{ij}$ and scale $\Delta S_{ij}$ are also computed by subtracting directly from corresponding parameters. Since spatial information is inferred on a sparse graph, it is likely for some objects to have no connections with others. In this case, we only decode detection results from their node embeddings. 

\begin{figure}[htbp]
    \centering
    \includegraphics[width=\linewidth]{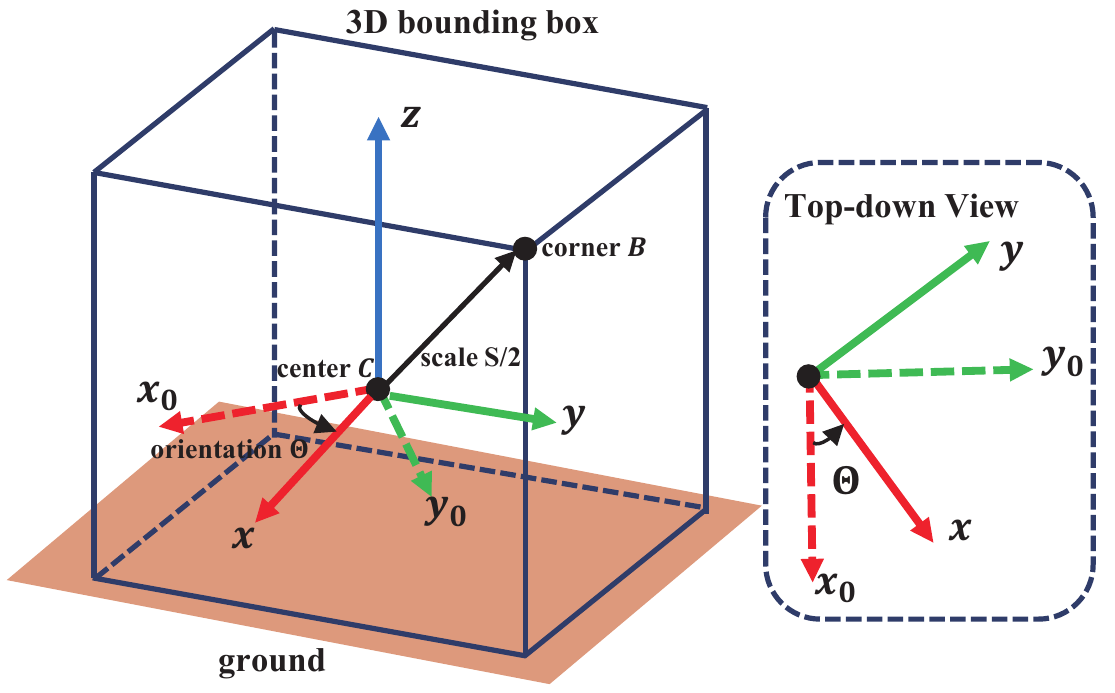}
    \caption{Illustration of the frame bound to 3D bounding box.}
    \label{figframe}
\end{figure}

Homogeneous transformation is introduced in this paper to better interpret spatial relationship and form our holistic estimate. As shown in Fig. \ref{figframe}, a 3D bounding box can be modeled in a coordinate system manner with its origin located at the cuboid center and its $X$-axis corresponding to $\Theta_i$. This way, this bounding box frame can be represented as a homogeneous matrix $^WT_i \in \mathbb{R}^{4 \times 4}$, which denotes the transformation with respect to the world coordinate system. 

According to the principle of homogeneous transformation, the spatial difference between node $i$'s frame $^WT_i$ and node $j$'s frame $^WT_j$ is defined as Eq. \ref{eqhomo}, where $^iT_j$ denotes the "difference" spatial transformation and $\Delta \Theta_{ij}, \Delta C_{ij}$ denotes the relative pose predicted by relative decoder and $R_z(\cdot)$ denotes the rotation matrix along $Z$-axis. Relative scale $\Delta S$ cannot be measured directly in this homogeneous manner, however, we can instead compute the translation between bounding box corners to aid bounding box scale prediction as Eq. \ref{eqscale}, where $^WB_i, ^WB_j$ denotes the bounding box $i$'s and $j$'s corners with respect to world frame.
\begin{equation}
\label{eqhomo}
\begin{aligned}
^WT_i = &
\begin{bmatrix}
R_z(\Theta_{i}) & C_{i} & \\
0 & 1 &
\end{bmatrix} \\
^iT_j = &
\begin{bmatrix}
R_z(\Delta \Theta_{ij}) & \Delta C_{ij} & \\
0 & 1 &
\end{bmatrix} \\
^W\hat{T}_j = & ^iT_j ^WT_i \\
\end{aligned}
\end{equation}

\begin{equation}
\begin{aligned}
\label{eqscale}
^WB_i = &
\begin{bmatrix}
R_z(\Theta_{i}) & C_{i}+S_{i}/2 & \\
0 & 1 &
\end{bmatrix} \\
^iB_j = &
\begin{bmatrix}
R_z(\Delta \Theta_{ij}) & \Delta C_{ij} + \Delta S_{ij}/2 & \\
0 & 1 &
\end{bmatrix} \\
^W\hat{B}_j = & {^iB_j} {^WB_i}
\end{aligned}
\end{equation}

Our final estimate of object-wise 3D bounding boxes incorporates independent prediction and relational inference results in a manner of weighted sum. Take a scene with two objects as an example. Assuming the independent parameters of the object $i$, $j$ and the relative parameters from $i$ to $j$ have been decoded as mentioned above, our task is to give the final estimate of the object $i$'s spatial parameters. In addition to the direct estimates provided by the object decoder, the spatial transformation from the relative decoder can also be fused with object $j$'s independent parameters to predict the object $i$'s bounding box according to the theory of Eq. \ref{eqhomo} and \ref{eqscale}. Thus the holistic prediction can be defined as Eq. \ref{eqholistic}.
\begin{equation}
\begin{aligned}
\label{eqholistic}
& ^W\tilde{T}_j = \alpha ^W\hat{T}_j + \beta \sum_{i=1}^{n} \overline{r}_{ij} {^kT_j} {^WT_i} \\
& ^W\tilde{B}_j = \alpha ^W\hat{B}_j + \beta \sum_{i=1}^{n} \overline{r}_{ij} {^kB_j} {^WB_i}
\end{aligned}
\end{equation}
where, $\alpha, \beta$ controls weight of independent and relative prediction, $\overline{r}_{ij}$ denotes the normalized relatedness score across connections and the final estimation of spatial parameters is given by $^W\tilde{T}_j, ^W\tilde{B}_j$. We can extract these variables conveniently from the homogeneous matrix if the specific object pose and scale parameters are needed (for example, for calculating loss function). 

\subsection{Loss Function}
For loss functions, to leverage the spatial transformation error between targets to boost supervised learning, we designed loss functions in both independent and relative predictions. We conclude the learning targets with both individual and relative losses as follows.

\subsubsection{Individual Loss}  
Spatial parameters $\delta, d, s, \theta$ in the camera system are predicted by the object decoder from the node embedding. It should be noted that camera-space parameters instead of world-system parameters are used for training here because they need accurate camera rotation angles to give confident prediction results, while camera-space parameters can be regressed under any task without additional requirements of other unknown parameters.

Since directly regressing absolute angles or length with L2 loss is error-prone \cite{huang2018cooperative, qi2018frustum}. We consider the prediction of orientation $\theta$, scale $s$ and centroid distance $d$ as a classification task among fixed bins and a regression task within the selected bin. And classification-regression loss $\mathcal{L}^{cls, reg}=\mathcal{L}^{cls}+\lambda \mathcal{L}^{reg}$ in \cite{qi2018frustum} is used to optimize these parameters, where $\lambda$ controls the weight of regression task, $\mathcal{L}^{cls}$ is calculated as cross-entropy loss and $\mathcal{L}^{reg}$ is calculated as MSE loss. In terms of 3D-2D re-projection error $\delta$, since it is calculated by pixel offset from the centre and only ranges from the 2D plane and does not suffer from dimension inconsistency, thus we use L2 loss to predict $\delta$. The total individual loss can be formulated as Eq. \ref{eqsingleloss}, where $\delta^*$ denotes ground-truth re-projection offset.
\begin{equation}
    \label{eqsingleloss}
    \mathcal{L}_{individual} = \mathcal{L}^{cls,reg}_{d} + \mathcal{L}^{cls,reg}_{s} + \mathcal{L}^{cls,reg}_{\theta} + \left \| \delta - \delta^* \right \|_2
\end{equation}

\subsubsection{Relative Loss} 
The spatial transformation parameters decoded from the relative decoder can aid supervised learning either as direct estimates or as a component of holistic estimates. For direct estimate loss, the ground-truth spatial transformation parameters $\Delta C_{ij}^*, \Delta S_{ij}^*, \Delta \Theta_{ij}^*$ can be calculated preliminarily, thus we can derive direct transformation loss based on L2 loss, as Eq. \ref{eqreldirect}. It should be noted that, unlike the individual loss mentioned above, L2 loss instead of classification-regression loss is used here because spatial transformation error is differentiable \cite{campa2009acc}.
\begin{equation}
\label{eqreldirect}
\mathcal{L}_{direct} = \left \| \Delta C - \Delta C^* \right \|_2 + \left \| \Delta S - \Delta S^* \right \|_2 + \left \| \Delta \Theta - \Delta \Theta^* \right \|_2 
\end{equation}
The relative spatial transformation parameters can also incorporate with independent parameters of other objects to form a holistic estimate as Eq. 2. Thus, the holistic loss can be defined as the error between ground-truth parameters $C^*, S^*, \Theta^*$ and their counterparts in holistic estimation. As shown in Eq. \ref{eqrelholistic}, $f_T$ and $f_B$ respectively map the corresponding homogeneous matrix into pose parameters $[\Theta, C]$ and scale parameter $S$. Based on the same considerations as $\mathcal{L}_{direct}$, L2 loss is also used here.
\begin{equation}
\label{eqrelholistic}
\mathcal{L}_{holistic} = \left \| f_T(^W\hat{T})  - [\Theta, C]^T \right \|_2 + \left \| f_B(^W\hat{B}) - S \right \|_2 
\end{equation}

\subsubsection{Corner Loss}
Since the coordinates of ground-truth bounding box corners can be figured out by Eq.\ref{eqholistic}, we can also use the difference between actual coordinates and their prediction as a part of learning loss. Direct regression of coordinates has been proved to be an effective way for transformations \cite{wang2018pixel2mesh, pan2019deep}, thus the corner loss is defined as the Eucliean distance between ground-truth and prediction corners as shown in Eq.\ref{eqcornerloss}. Note that the corner losses from object-wise and relative prediction are separately computed because incorporation will lose some information for supervised learning. 
\begin{equation}
\label{eqcornerloss}
\mathcal{L}_{corner} = \left \| ^WB^* - ^WB \right \|_2 + \left \| ^WB^* - ^W\hat{B} \right \|_2
\end{equation}
where $^WB$ and $^W\hat{B}$ denotes bounding box corners computed as Eq. \ref{eqholistic} respectively.

\subsubsection{Physical Violation Loss}
During object detection, bounding boxes of different objects should obey the rule of the physical world, which is not to have intersections. Some objects overlap in real-world environments (for example, pillows on the bed). However, here, we only focus on typical indoor furniture such as beds, tables, chairs etc. and leave out those small, movable objects. We employ a similar physical violation loss as \cite{huang2018cooperative} as Eq. \ref{eqphysical}. By introducing the geometric constraint in the physical world, we can penalize those bounding boxes with collisions and the greater the overlapping is, the more penalty there will be. This way, the prediction results in a more natural scene understanding.
\begin{equation}
\label{eqphysical}
\begin{aligned}
\mathcal{L}_{physical} =  \sum_{i=1}^{n} \sum_{j=1}^{n} & {(ReLU(Max(^WB_j)-Max(^WB_i)))} \\ 
+ & {ReLU(Min(^WB_j) - Min(^WB_i))} \\
\end{aligned}
\end{equation}
where $ReLU(\dot)$ is the activation function, $Max(\dot)/Min(\dot)$ function takes a 3D bounding box as the input and outputs the maximum or minimum value along three axis.

The entire loss function can be represented as Eq. \ref{eqall}, where $\lambda_1, \lambda_2, \lambda_3$ are the weights for the corresponding loss. In our implementation, these three weights are set to 0.75, 0.6, 0.8 respectively.
\begin{equation}
\label{eqall}
\begin{aligned}
    \mathbb{L} = & \mathbb{L}_{individual} \\ & + \lambda_1 * (\mathbb{L}_{direct} + \mathbb{L}_{holistic}) \\
    & + \lambda_2 * \mathbb{L}_{corner} \\ 
    & + \lambda_3 * \mathbb{L}_{physical} \\
\end{aligned}
\end{equation}

\section{Experiment \label{exp}}
\subsection{Experimental Setup}
In this paper, the input to the detection system is always a 2D image along with the detected 2D bounding boxes obtained from Faster R-CNN\cite{ren2015faster} object detector. We report detection results on both tasks: 1) 3D object detection, which outputs 3D box predictions for each 2D box. 2) semantic reconstruction, which outputs the 3D layout estimation, object-wise 3D detection and object-centric meshes for each object. We summarize our experimental settings in Table \ref{taboverview}.
\begin{table}[!htbp]
	\caption{The experimental configurations.}
	  \label{taboverview}
        \resizebox{\columnwidth}{!}{
	\begin{tabular}{c|c|c|c}
		\toprule[1pt]
		\textbf{Section} & \textbf{Task}   & \textbf{Input}   & \textbf{Output}   \\ \midrule
		Sec. \ref{section:exp3d}    & \begin{tabular}[c]{@{}c@{}}3D object \\ detection\end{tabular}    & \begin{tabular}[c]{@{}c@{}}2D images,\\ detection boxes,\\ camera orientation \end{tabular}  & 3D box and pose \\ \midrule
		Sec. \ref{section:expsec}  & \begin{tabular}[c]{@{}c@{}}semantic\\ reconstruction\end{tabular} & \begin{tabular}[c]{@{}c@{}}2D images, \\ detection boxes\end{tabular} & \begin{tabular}[c]{@{}c@{}}3D box and Pose,\\ scene layout, \\ object mesh\end{tabular} \\ \bottomrule[1pt]
	\end{tabular}}
\end{table}

\subsubsection{Datasets:} We use the SUN RGB-D \cite{song2015sun} dataset with the official training and testing split and NYU-37 \cite{silberman2012indoor} object labels for evaluation on 3D object detection. The SUN RGB-D dataset consists of 10,335 dense annotated real indoor images with coarse point cloud, accurate 3D object bounding boxes and labeled 3D layout. Pix3D \cite{sun2018pix3d} dataset is also used in semantic reconstruction task, but it is only for mesh generation and joint training of semantic reconstruction. This dataset provides 395 furniture models with nine categories aligned with 10,069 images, The training and testing split is kept inline with \cite{gkioxari2019mesh}, and we refer readers to \cite{nie2020total3dunderstanding} for the details about the mapping from NYU-37 label to Pix3D label.

\subsubsection{Metrics:} Our results are mainly measured on 3D object detection metrics in \cite{tulsiani2018factoring}, including the average precision (AP), translation, rotation and scale. In our implementation, a 3D IoU threshold to determine a prediction as a true positive is set to 0.15 for both tasks.
\begin{table*}[bp]
	\caption{Comparisons of object pose prediction on on 3D Object Detection and Semantic Reconstruction. The difference values of distance to centroid (distance), orientation angle (rotation) and volume scaling coefficient (scale) are reported.}
	   \label{tabpose}
	\resizebox{\textwidth}{!}{
	\begin{tabular}{c|l|ccc|ccc|ccc}
		\toprule[1pt]
		\multirow{3}{*}{Task}  & \multicolumn{1}{c|}{\multirow{3}{*}{Method}} & \multicolumn{3}{c|}{Translation(meters)}  & \multicolumn{3}{c|}{Rotation(degrees)}  & \multicolumn{3}{c}{Scale}  \\ 
		& \multicolumn{1}{c|}{}  & Median  & Mean  & (Err$\le$0.5m)\%   & Median  & Mean  & (Err$\le$30\degree)\% & Median  & Mean & (Err$\le$0.2)\%    \\
		& \multicolumn{1}{c|}{} & \multicolumn{2}{c}{(lower is better)} & (higher is better) & \multicolumn{2}{c}{(lower is better)} & (higher is better) & \multicolumn{2}{c}{(lower is better)} & (higher is better) \\ \midrule
		\multirow{4}{*}{\begin{tabular}[c]{@{}c@{}}3D Object\\ Detection\end{tabular}} 
		& GCN         & 0.59 & 0.70 & 43.2 & 18.5 & 48.6 & 61.5 & 0.36 & 0.39 & 39.9 \\
		& InterNet    & 0.56 & 0.69 & 44.6 & 18.3 & 48.2 & 61.8 & 0.35 & 0.37 & 41.2 \\
		& Total3D     & 0.52 & 0.65 & 49.2 & 17.6 & 45.1 & 64.1 & 0.28 & 0.29 & 42.1 \\
		& \textbf{Ours} & \textbf{0.50} & \textbf{0.59} & \textbf{55.4} & \textbf{13.5} & \textbf{44.3} & \textbf{64.3} & \textbf{0.25} & \textbf{0.27} & \textbf{42.8}      \\ \midrule
		\multirow{4}{*}{\begin{tabular}[c]{@{}c@{}}Semantic\\ Reconstruction\end{tabular}}              
		& GCN         & 0.55 & 0.66 & 45.7 & 16.8 & 46.1 & 63.8 & 0.33 & 0.34 & 40.5 \\
		& InterNet    & 0.52 & 0.64 & 46.9 & 15.2 & 43.8 & 65.5 & 0.28 & 0.29 & 42.6 \\
		& Total3D     & 0.48 & 0.61 & 51.8 & 14.4 & 43.7 & 66.5 & 0.22 & 0.26 & \textbf{43.7}  \\  
		& \textbf{Ours} & \textbf{0.46} & \textbf{0.55} & \textbf{57.3} & \textbf{10.2} & \textbf{42.6} & \textbf{66.8} & \textbf{0.21} & \textbf{0.24} & 43.5 \\ 
		\bottomrule[1pt]
	\end{tabular}}
\end{table*}

\begin{table*}[bp]
	\caption{Comparison of average precision (AP) on 3D Object Detection and Semantic Reconstruction (higher is better).}
	   \label{tabmap}
	\resizebox{\textwidth}{!}{
		\begin{tabular}{c|l|c|c|c|c|c|c|c|c|c|c|c}
		\toprule[1pt]
		Task  & \multicolumn{1}{c|}{Method} & bed   & chair & sofa  & table & desk  & dresser & nightstand & sink  
            & cabinet & lamp & \multicolumn{1}{c}{mAP} \\ \midrule
		\multirow{4}{*}{\begin{tabular}[c]{@{}c@{}}3D Object\\ Detection \end{tabular}} 
		
            & GCN      & 58.95 & 13.55 & 32.28 & 28.68 & 19.21 & 12.31  & 5.99 & 12.01 & 10.62   & 2.25 & 19.58 \\  
		& InterNet & 60.58 & 15.02 & 35.35 & 32.96 & 20.25 & 15.01  & 6.33 & 12.64 & 11.25   & 2.66 & 21.20 \\ 
		& Total3D  & 59.03 & 15.98 & 43.95 & 35.28 & 23.65 & 19.20  & 6.87 & 14.40 & 11.39   & 3.46 & 23.32 \\ 
            & \textbf{Ours} & \textbf{60.89} & \textbf{16.86} & \textbf{46.25} & \textbf{35.52} & \textbf{25.36} & \textbf{21.56} & \textbf{8.25}  & \textbf{16.89} & \textbf{13.61} & \textbf{4.54} & \textbf{24.97} \\ 
            
            \midrule
  
		\multirow{8}{*}{\begin{tabular}[c]{@{}c@{}}Semantic \\ Reconstruction\end{tabular}}
            & 3DGP     & 5.62  & 2.31  & 3.24  & 1.23  & -     & -      & -    & -     & -       & -    & -   \\
            & HoPR     & 58.29 & 13.56 & 28.37 & 12.12 & 4.79  & 13.71  & 8.80 & 2.18  & 0.48    & 2.41 & 14.47 \\
            & CooP     & 57.71 & 15.21 & 36.67 & 31.16 & 19.90 & 15.98  & 11.36 & 15.95  & 10.47    & 3.28 & 21.77 \\ 
		  & GCN     & 57.95 & 15.55 & 36.28 & 32.68 & 
            22.21 & 18.31   & 10.21      & 15.01 & 11.62   & 3.25 & 22.31  \\
		  & InterNet & 61.24 & 16.82 & 40.68 & 35.96 & 
            28.25 & 20.01   & 13.33      & 16.64 & 13.25   & 4.66 & 25.08  \\
		  & Total3D   & 60.65 & 17.55 & 44.90 & 36.48 & 
            27.65 & 21.19   & 17.01      & 18.50 & 14.50   & 5.04 & 26.38 \\
            & \textbf{Ours} & \textbf{63.35} & \textbf{19.86} & \textbf{49.25} & \textbf{37.52} & \textbf{28.55} & \textbf{24.18}   & \textbf{19.33}      & \textbf{21.89} & \textbf{17.61}   & \textbf{5.89} & \textbf{28.74} \\
            & \textcolor[rgb]{1,0,0}{Im3D}     & \textcolor[rgb]{1,0,0}{89.32} & \textcolor[rgb]{1,0,0}{35.14} & \textcolor[rgb]{1,0,0}{69.10} & \textcolor[rgb]{1,0,0}{57.37} & \textcolor[rgb]{1,0,0}{49.03} & \textcolor[rgb]{1,0,0}{29.27} & \textcolor[rgb]{1,0,0}{41.34} & \textcolor[rgb]{1,0,0}{33.81} & \textcolor[rgb]{1,0,0}{33.93} & \textcolor[rgb]{1,0,0}{11.90} & \textcolor[rgb]{1,0,0}{45.21} \\ 

		\bottomrule[1pt]
	\end{tabular}}
\end{table*}

\subsubsection{Baselines:} We use the following baseline methods in our experiments:
\begin{itemize}[leftmargin=10pt]
	\item \emph{GCN}: We use the object encoder in Sec. \ref{graph} to obtain object-wise embeddings $o_i$ and then adopt a two-layer \emph{GCN} \cite{kipf2016semi} to perform implicit relational reasoning. These final embeddings are then fed into the decoder for the prediction of 3D object pose.
	
	\item \emph{InteractionNet}: \emph{InteractionNet} \cite{battaglia2016interaction} serves as an alternate way to reason implicitly over the dense object embeddings. We use a learned MLP to obtain the \emph{effect} embedding $e_{AB}$ for each ordered tuple ($o_A, o_B$) and then aggregate these embeddings by $o_A+max_{B}(e_{AB})$ and update them. The final embeddings are used for per-object predictions.
	
	\item \emph{Total3D}: We use the method from \cite{nie2020total3dunderstanding} to perform relational reasoning with attention mechanism. We use the ResNet encoder to obtain appearance features and the relation module to calculate per-object relational features. These are then added to the target in an element-wise manner to form the final target feature and regress each set of box parameters.
\end{itemize}

\subsection{3D Object Detection Evaluation \label{section:exp3d}}
We first analyze the performance of all baseline methods in the setting of single image 3D object detection. For this task, given a scene image with detected 2D bounding boxes, we train all the baselines with the batch size of 16, total epochs of 100 and the learning rate of 1e-3. Each baseline is trained to output the final object embeddings, and the decoder predicts the 3D pose of the objects. 

During testing, all baselines predict the 3D posture, including translation, rotation and scale for each object. Detection AP is calculated under the 3D IoU threshold, while the translation, orientation, and scale error are calculated as the absolute difference. As shown in Table \ref{tabpose} and \ref{tabmap}, we report the mean error of pose and the mAP of 10 selected categories across the different baselines. 

\subsubsection{Qualitative Results} 
\begin{figure*}[bp]
	\begin{minipage}{0.19\linewidth}
	\vspace{3pt}
	\includegraphics[width=0.95\textwidth, height=0.1\textheight]{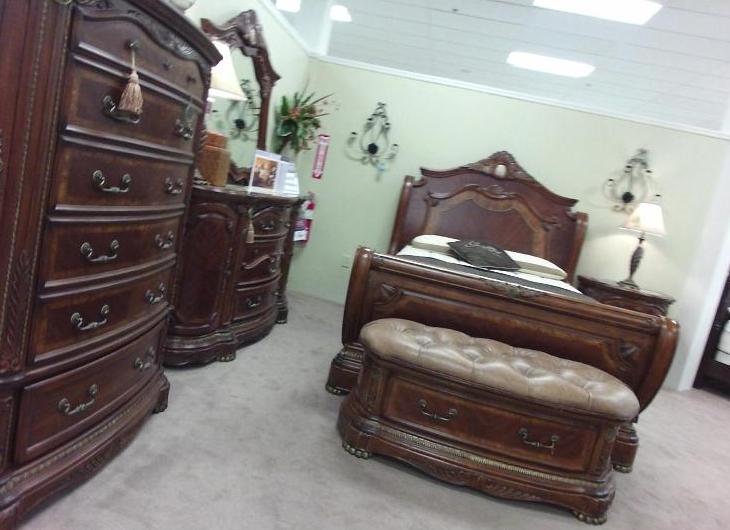}
	\vspace{0.5pt}
	\includegraphics[width=0.95\textwidth, height=0.1\textheight]{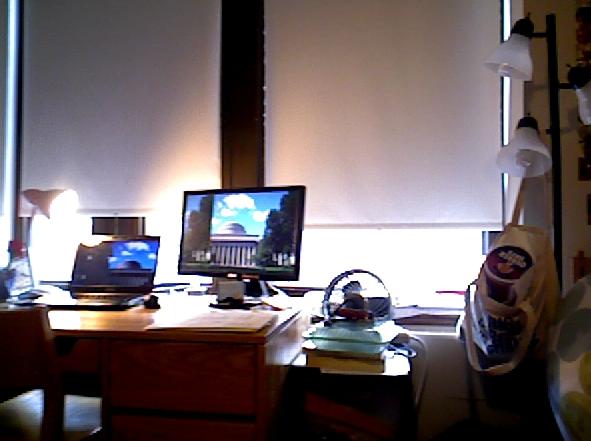}
	\vspace{3pt}
	\includegraphics[width=0.95\textwidth, height=0.1\textheight]{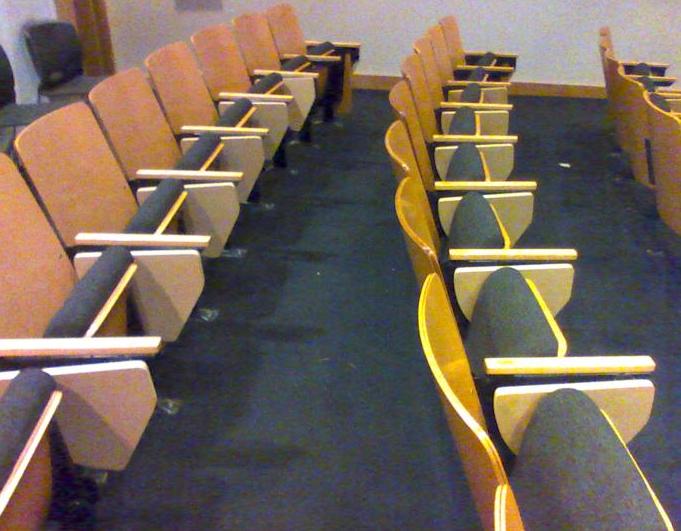}
	\centerline{(a)}
	\end{minipage}
	\begin{minipage}{0.15\linewidth}
	\vspace{3pt}
	\includegraphics[width=0.95\textwidth, height=0.1\textheight]{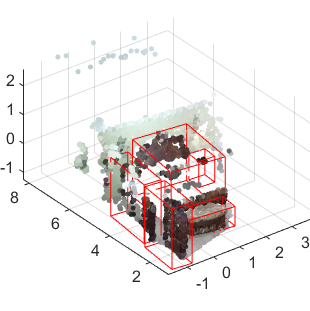}
	\vspace{0.5pt}
	\includegraphics[width=0.95\textwidth, height=0.1\textheight]{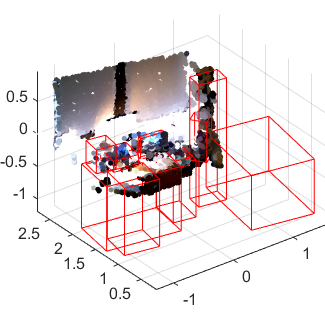}
	\vspace{3pt}
	\includegraphics[width=0.95\textwidth, height=0.1\textheight]{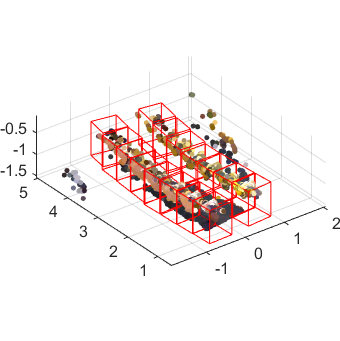}
	\centerline{(b)}
	\end{minipage}
	\begin{minipage}{0.15\linewidth}
	\vspace{3pt}
	\includegraphics[width=0.95\textwidth, height=0.1\textheight]{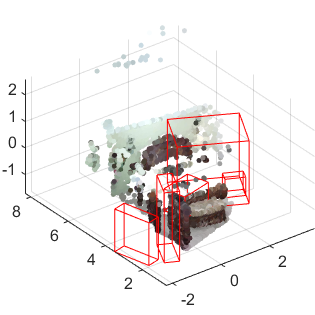}
	\vspace{0.5pt}
	\includegraphics[width=0.95\textwidth, height=0.1\textheight]{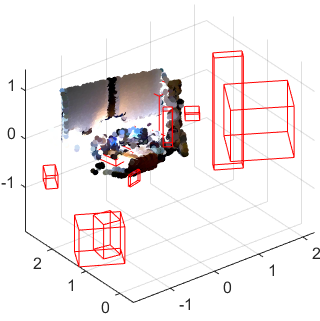}
	\vspace{3pt}
	\includegraphics[width=0.95\textwidth, height=0.1\textheight]{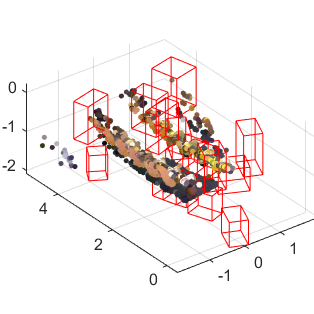}
	\centerline{(c)}
	\end{minipage}
	\begin{minipage}{0.15\linewidth}
	\vspace{3pt}
	\includegraphics[width=0.95\textwidth, height=0.1\textheight]{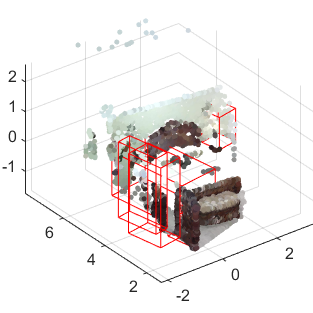}
	\vspace{0.5pt}
	\includegraphics[width=0.95\textwidth, height=0.1\textheight]{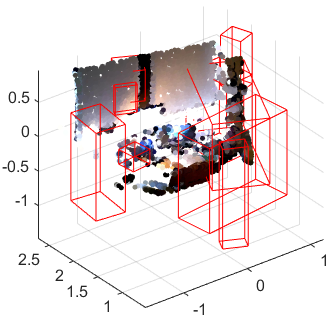}
	\vspace{3pt}
	\includegraphics[width=0.95\textwidth, height=0.1\textheight]{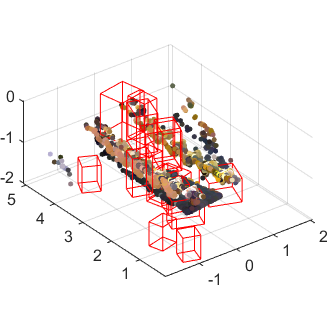}
	\centerline{(d)}
	\end{minipage}
	\begin{minipage}{0.15\linewidth}
	\vspace{3pt}	
	\includegraphics[width=0.95\textwidth, height=0.1\textheight]{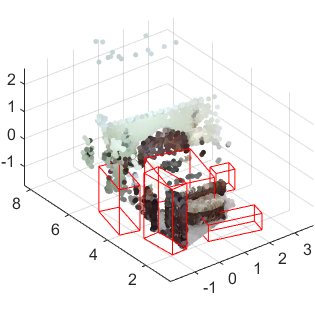}
	\vspace{0.5pt}
	\includegraphics[width=0.95\textwidth, height=0.1\textheight]{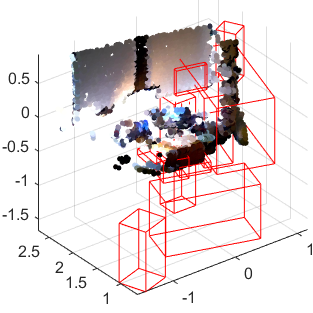}
	\vspace{3pt}
	\includegraphics[width=0.95\textwidth, height=0.1\textheight]{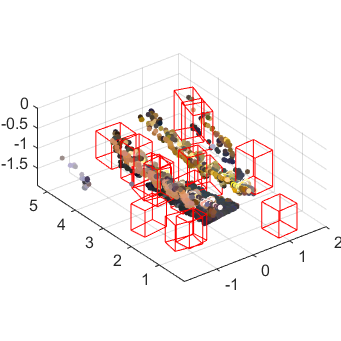}
	\centerline{(e)}
	\end{minipage}
	\begin{minipage}{0.15\linewidth}
	\vspace{3pt}	
	\includegraphics[width=0.95\textwidth, height=0.1\textheight]{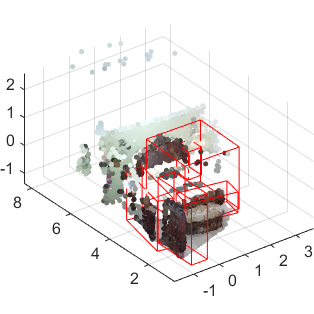}
	\vspace{0.5pt}
	\includegraphics[width=0.95\textwidth, height=0.1\textheight]{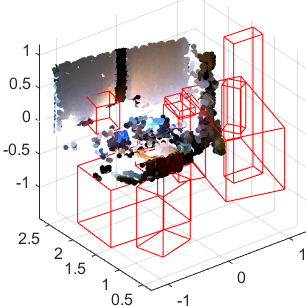}
	\vspace{3pt}
	\includegraphics[width=0.95\textwidth, height=0.1\textheight]{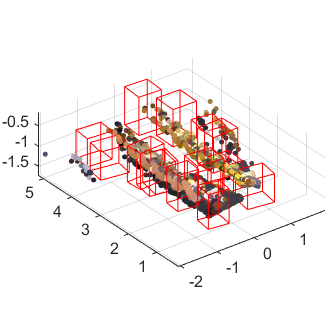}
	\centerline{(f)}
	\end{minipage}
	\caption{Visualization of 3D object detection results.From left to right:(a)Input images, (b) Ground-truth and results from (c) GCN \cite{kipf2016semi}, (d) InteractionNet \cite{battaglia2016interaction}, (e) Total3D \cite{nie2020total3dunderstanding}, (f) Ours.}
	 \label{expfig3d}
\end{figure*}

In Fig. \ref{expfig3d}, we use the official SUN RGB-D MATLAB toolbox for visualization and show a few results of the studied baseline methods. Compared to \emph{Total3D}, our results exhibit more uniformly oriented objects as well as better stability and consistency in complex and clutter environments (Row No.3). Besides, it can be seen from Row No.2 that our results show less physical violation with better ability to predict objects at the edge of the image or even partially behind the imaging plane. We can conclude from the visualization that \emph{Total3D} and our method perform better than \emph{GCN} and \emph{InteractionNet} with more accurate object location and orientation. 

\begin{figure*}[htbp]
	\begin{minipage}{0.19\linewidth}
	\vspace{3pt}
	\includegraphics[width=0.95\textwidth, height=0.1\textheight]{det_results/imgs/img1.jpg}
	\vspace{0.5pt}
	\includegraphics[width=0.95\textwidth, height=0.1\textheight]{det_results/imgs/img5.jpg}
	\vspace{3pt}
	\includegraphics[width=0.95\textwidth, height=0.1\textheight]{det_results/imgs/img6.jpg}
	\centerline{(a)}
	\end{minipage}
	\begin{minipage}{0.19\linewidth}
	\vspace{3pt}
	\includegraphics[width=0.95\textwidth, height=0.1\textheight]{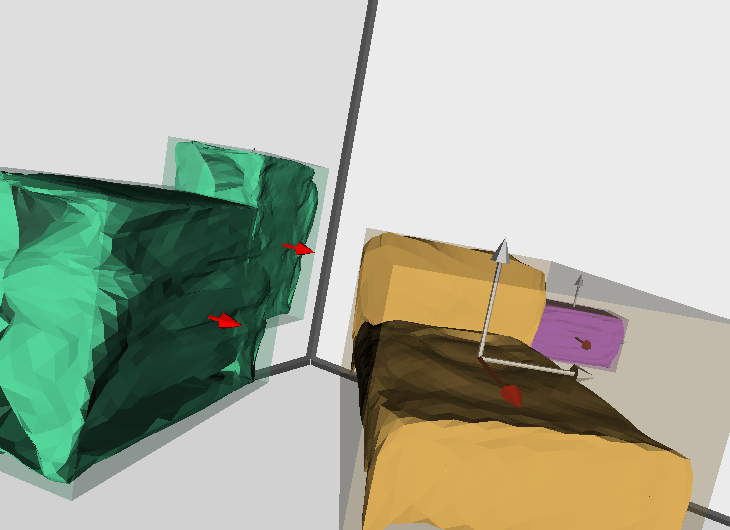}
	\vspace{0.5pt}
	\includegraphics[width=0.95\textwidth, height=0.1\textheight]{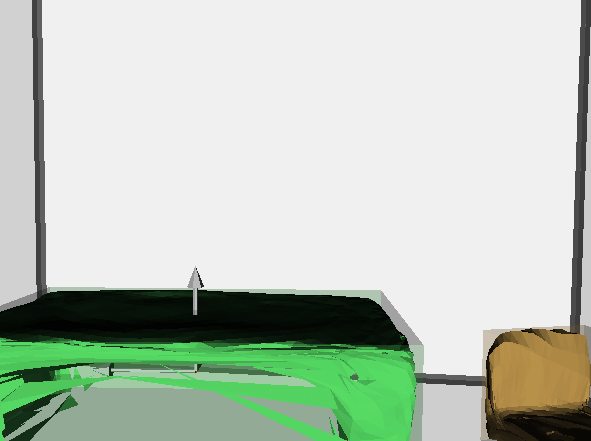}
	\vspace{3pt}
	\includegraphics[width=0.95\textwidth, height=0.1\textheight]{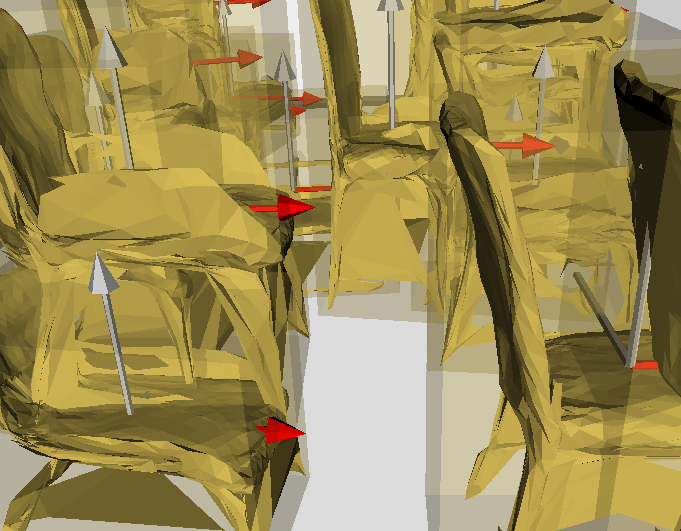}
	\centerline{(b)}
	\end{minipage}
    \begin{minipage}{0.19\linewidth}
   	\vspace{3pt}
   	\includegraphics[width=0.95\textwidth, height=0.1\textheight]{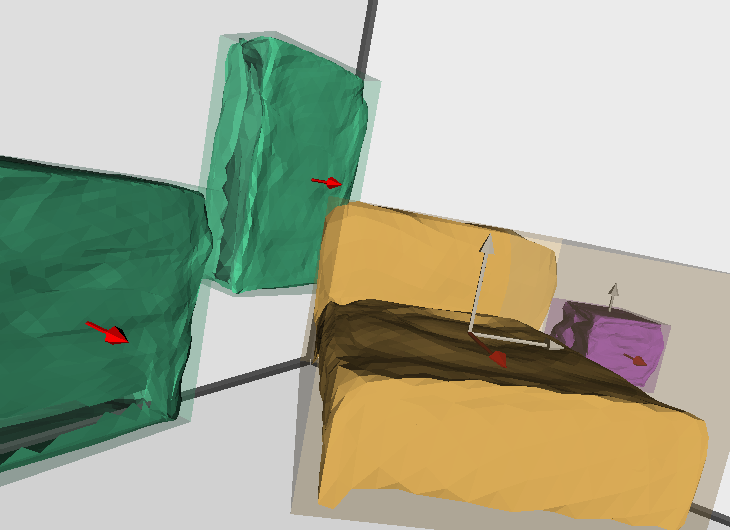}
    \vspace{0.5pt}
   	\includegraphics[width=0.95\textwidth, height=0.1\textheight]{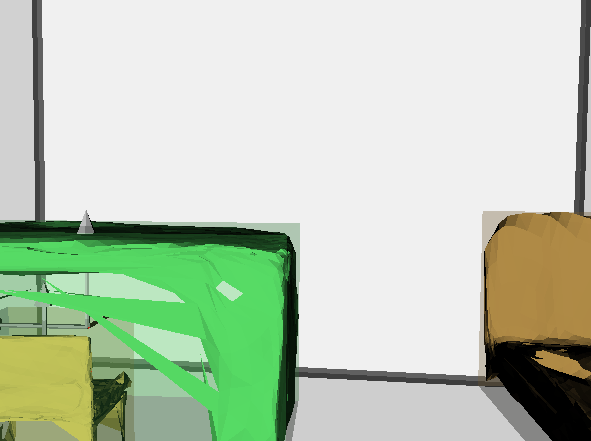}
   	\vspace{3pt}
   	\includegraphics[width=0.95\textwidth, height=0.1\textheight]{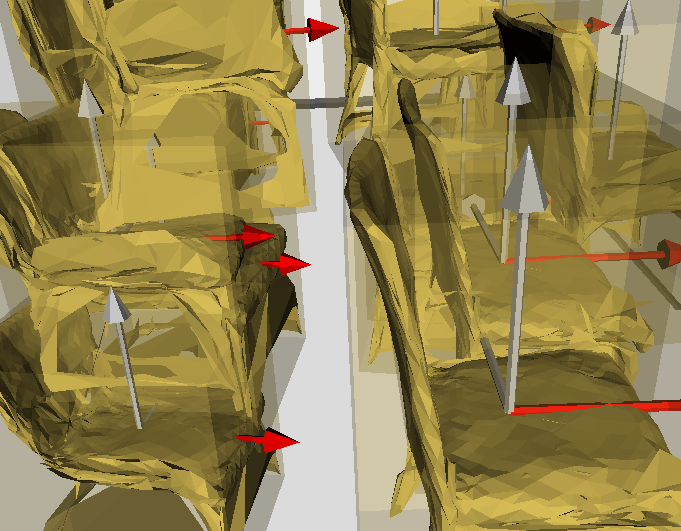}
   	\centerline{(c)}
    \end{minipage}
	\begin{minipage}{0.19\linewidth}
	\vspace{3pt}	
	\includegraphics[width=0.95\textwidth, height=0.1\textheight]{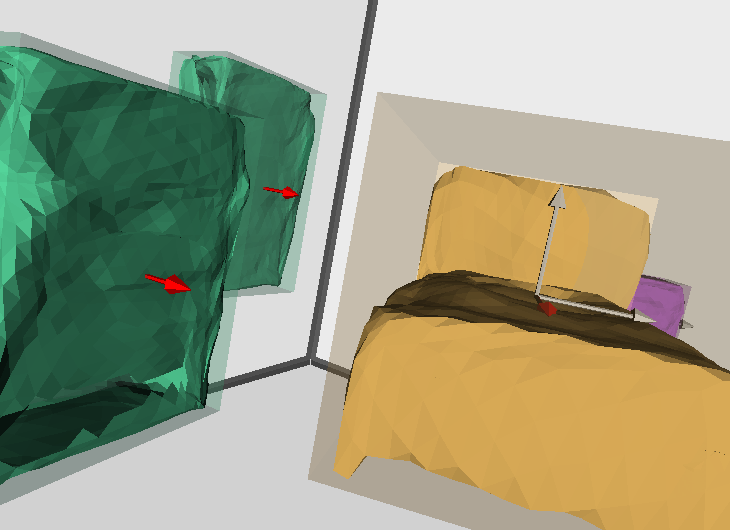}
	\vspace{0.5pt}
	\includegraphics[width=0.95\textwidth, height=0.1\textheight]{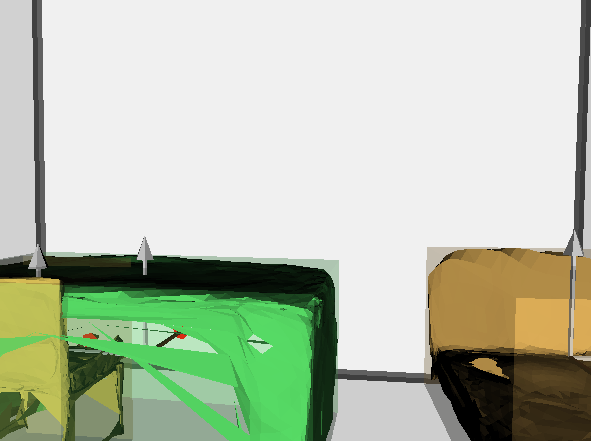}
	\vspace{3pt}
	\includegraphics[width=0.95\textwidth, height=0.1\textheight]{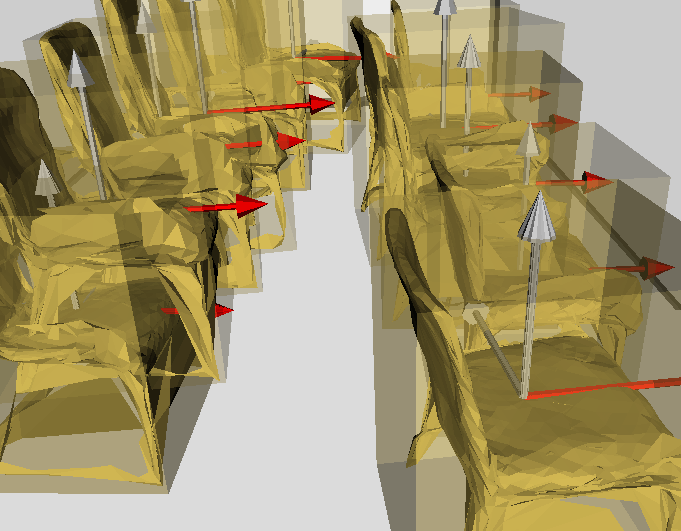}
	\centerline{(d)}
	\end{minipage}
    \begin{minipage}{0.19\linewidth}
   	\vspace{3pt}	
   	\includegraphics[width=0.95\textwidth, height=0.1\textheight]{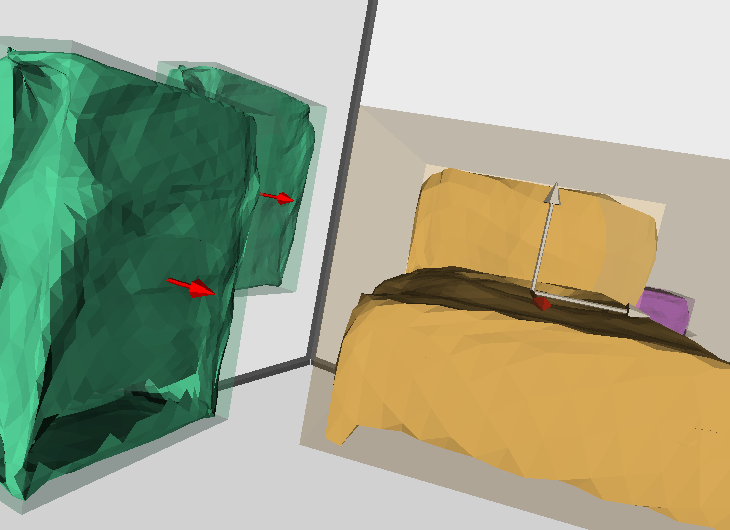}
   	\vspace{0.5pt}
    \includegraphics[width=0.95\textwidth, height=0.1\textheight]{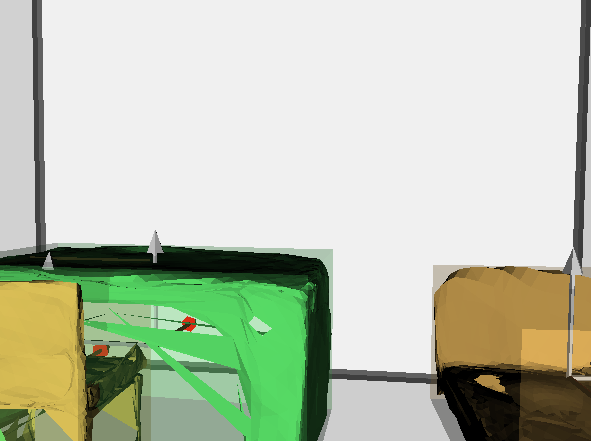}
    \vspace{3pt}
   	\includegraphics[width=0.95\textwidth, height=0.1\textheight]{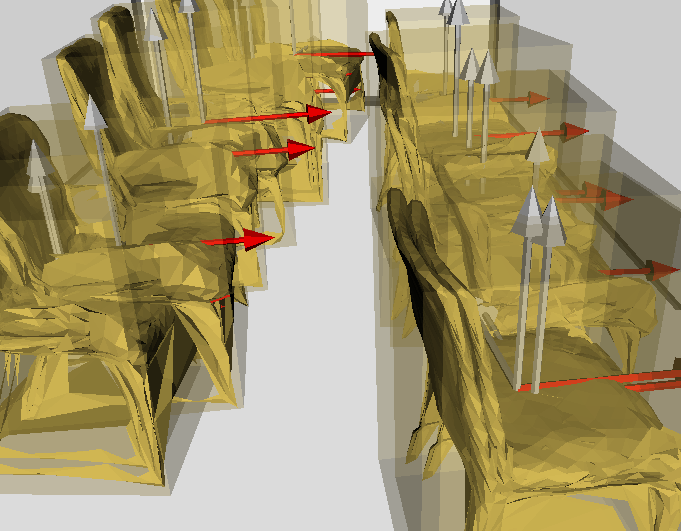}
   	\centerline{(e)}
    \end{minipage}
	\caption{Visualization of semantic reconstruction results. From left to right:(a)Input images and results from (b) GCN \cite{kipf2016semi}, (c) InteractionNet \cite{battaglia2016interaction}, (d) Total3D \cite{nie2020total3dunderstanding}, (e) Ours.}
	 \label{expfigsec}
\end{figure*}

\subsubsection{Quantitative Results} 
Both \emph{GCN} and \emph{InteractionNet} implicitly reason over dense pairwise connections, while \emph{Total3D} incorporates the pairwise relations with attention mechanism. From the detection results illustrated in Table \ref{tabpose} and \ref{tabmap}, \emph{Total3D} boosts significantly over the other two implicit reasoning models. Our approach goes further to explicitly model the spatial difference, which enables us to gain by a small margin over other methods.

\subsection{Semantic Reconstruction Evaluation \label{section:expsec}}
To extend our graph model in other scene understanding applications and evaluate its robustness, we also combine our 3D object detection model with layout estimation and mesh reconstruction methods to perform the holistic semantic reconstruction. We keep inline with \cite{nie2020total3dunderstanding} and use the Mesh Generation Network (MGN) to reconstruct the object-centric object meshes. Layout Estimation Network (LEN) predicts the camera pose $R(\beta, \gamma)$ and room layout $(C, s^l, \theta^l)$ in the world system.
We present the architecture of the Layout Estimation Network (Table \ref{lentab}), Mesh Generation Network (Table \ref{mgntab}, note that $d_c=9$ and $N_e=$ in our experiment) used in semantic reconstruction task as follows. The configuration of MGN and LEN are kept consistent with \cite{nie2020total3dunderstanding}.
\begin{table}[!ht]
	\caption{Layout Estimation Network architecture.}
	\label{lentab}
	\resizebox{0.48\textwidth}{!}{
		\begin{tabular}{|c|c|c|c|}
			\hline 
			Index & Inputs & Operation  & Output shape \\ \hline \hline
			(1)   & Input & Scene image & 3$\times$ 256$\times$ 256 \\
			(2)   & (1)    & ResNet-34                  & 2048     \\
			(3)   & (2)    & FC(1024-d)+ReLU+Dropout+FC & $\beta$  \\
			(4)   & (2)    & FC(1024-d)+ReLU+Dropout+FC & $\gamma$ \\
			(5)   & (2)    & FC+ReLU+Dropout            & 2048  \\
			(6)   & (5)    & FC(1024-d)+ReLU+Dropout+FC & $C$   \\
			(7)   & (5)    & FC(1024-d)+ReLU+Dropout+FC & $s^l$ \\
			(8)   & (5)    & FC(1024-d)+ReLU+Dropout+FC & $\theta^l$ \\ \hline
	\end{tabular}}
\end{table}

\begin{table}[!h]
	\caption{Mesh Generation Network architecture.}
	  \label{mgntab}
	\resizebox{0.5\textwidth}{!}{
		\begin{tabular}{|c|c|c|c|}
			\hline 
			Index & Inputs & Operation & Output shape \\ \hline \hline
			(1)   & Input  & Object image & 3$\times$ 256$\times$ 256 \\
			(2)   & Input  & Object class code & $d_c$ \\
			(3)   & Input  & Template Sphere & 3$\times$2562 \\
			(4)   & (1)    & ResNet-18 & 1024 \\
			(5)   & (2),(4) & Concatenate & 1024+$d_c$ \\
			(6)   & (5)     & Repeat & (1024+$d_c$)$\times$ 2562 \\
			(7)   & (3),(6) & Concatenate & (1024+$d_c$+3)$\times$ 2562 \\
			(8)   & (7)     & AtlasNet decoder \cite{groueix2018papier} & 3$\times$2562  \\ 
			(9)   & (3),(8) & Element-wise sum & 3$\times$2562  \\ 
			(10)   & (9)    & Sample points & 3$\times N_e$ \\ 
			(11)   & (5)    & Repeat & (1024+$d_c$)$\times N_e$ \\ 
			(12)   & (10),(11) & Concatenate & (1024+$d_c$+3)$\times N_e$ \\ 
			(13)   & (12)    & Edge classifier & 1$\times N_e$ \\ 
			(14)   & (13)    & Threshold & 1$\times N_e$(Mesh topology) \\ 
			(15)   & (6),(9) & Concatenate & (1024+$d_c$+3)$\times$2562 \\ 
			(16)   & (15)    & AtlasNet decoder \cite{groueix2018papier} & 3$\times$2562 \\ 
			(17)   & (9),(16) & Element-wise sum & 3$\times$2562(Mesh points) \\ \hline
	\end{tabular}}
\end{table}

The training process is divided into 1) pre-training of Object Detection Network (ODN), LEN and MGN and 2) joint training of the entire model. We adopt the trained detection network in Sec. \ref{section:exp3d} as the pre-trained ODN. For the pre-training of LEN and MGN, we adopt a similar training scheme as Sec. \ref{section:exp3d}. We then combine all three individual modules to perform fine-tuning with the joint losses used in \cite{nie2020total3dunderstanding}. Here we integrate the scene image, the scene point cloud and $j$ object images into a hierarchical batch during training. During testing, all baseline methods only predict the object-wise 3D pose in the camera system. LEN and MGN predict the camera pose, layout bounding box and object-centric meshes respectively. We report pose error in Table \ref{tabpose} and object detection mAP in Table \ref{tabmap}. We also show the results of \emph{3DGP} \cite{choi2013understanding}, \emph{HoPR} \cite{huang2018holistic}, \emph{Coop} \cite{huang2018cooperative} and the SOTA method \emph{Im3D} \cite{zhang2021holistic} for a comprehensive comparison. We report the results of \emph{Coop} based on the model trained on the NYU-37 object labels for a fair comparison. Note that these three methods are not based on graphs, while \emph{GCN}, \emph{InterNet} and \emph{Im3D} are graph-based methods.

\subsubsection{Qualitative Results} 
In Fig. \ref{expfigsec}, we use VTK to visualize the semantic reconstruction, including object meshes and the holistic room layout. We observe that \emph{Total3D} and our method tends to output a similar and strictly ordered pose and layout while the other two exhibit severe physical violation.

\subsubsection{Quantitative Results} 
As shown in Table \ref{tabpose} and \ref{tabmap}, the detection metrics on semantic reconstruction tend to have a similar trend as Sec. \ref{section:exp3d}, where our method consistently outperforms other baselines. It should be noted that the 3D pose error in camera coordinate from the baseline methods will be further passed to other modules, making the results more error-prone. The consistent improvement on mAP and pose prediction also indicates our proposed method's stability and effectiveness for other scene understanding tasks.

\subsection{Ablation Analysis}
Our experiments demonstrate the advantages of modeling sparse relation and relative transformation for 3D object detection. To better understand the effect of each module on the final performance, we also test our model with three architecture variations:
\begin{itemize}[leftmargin=10pt]
	\item \textbf{$C_0$:} dense graph without relatedness score and relative loss (Baseline).
	\item \textbf{$C_1$:} Baseline + relatedness matrix
	\item \textbf{$C_2$:} Baseline + relative loss
	\item \textbf{Full:} Baseline + relatedness matrix + relative loss
\end{itemize}

\subsubsection{Effect of relatedness score.} Our method uses a relatedness score matrix to control the scene graph's dynamic pruning process. In Table \ref{tab3}, we study the effect of the relatedness score matrix and show our method's results with the fully connected graph in row 1 and 2. 

$C_0$ v.s. $C_1$ and $C_2$ v.s. \textbf{Full} shows that relatedness score helps to improve 3D object detection, which significantly reduces the rotation and scale error in 3D pose prediction.

\subsubsection{Effect of relative transformation reasoning.} Our graph model iteratively updates relative transformation prediction through the edge message update scheme, incorporating the relative loss to perform supervised learning on pairwise transformation. Table \ref{tab3} shows the results with relative loss in row 2 and 4. 

$C_0$ v.s.$C_2$ and $C_1$ v.s. \textbf{Full} shows that relative transformation prediction combined with relative loss consistently improves object detection metrics significantly either on dense graph or sparse graph.

\begin{table}[!htbp]
	\caption{Ablation analysis on 3D object detection.}
	  \label{tab3}
	\begin{tabular}{c|c|c|c|c}
		\toprule[1pt]
		Method & \begin{tabular}[c]{@{}c@{}}Translation\\ (meters)\end{tabular} & \begin{tabular}[c]{@{}c@{}}Rotation\\ (degrees)\end{tabular} & \begin{tabular}[c]{@{}c@{}}Scale \end{tabular} & \begin{tabular}[c]{@{}c@{}}3D Objects\\ (mAP) \end{tabular} \\ \midrule
		$C_0$ & 0.67 & 48.2 & 0.41 & 19.80 \\
		$C_1$ & 0.65 & 45.6 & 0.37 & 21.56 \\
		$C_2$ & \textbf{0.57} & 44.3 & 0.32 & 23.12 \\ \midrule
		\textbf{Full} & 0.59 & \textbf{43.6} & \textbf{0.2}7 & \textbf{24.97} \\
		\bottomrule[1pt]
	\end{tabular}
\end{table}

\subsection{Computational Cost Analysis}
We further analyze the number of parameters (Params) and floating point operations per second (FLOPS) of our algorithm to validate the computational efficiency of our sparse graph network. As shown in Table \ref{tabparams}, the computational cost of \emph{Total3D} \cite{nie2020total3dunderstanding} and \emph{Implicit3D} \cite{zhang2021holistic} are also listed here for comparison with ours because these three methods share the similar object detector architecture while follows distinct computation steps. Only Params and FLOPS of the object detector are presented here since our method is primarily designed for the single-image 3D detection task. 

\begin{table}[!htbp]
    \centering
    \caption{Comparison of Number of Parameters and FLOPS on Different Methods}
    \label{tabparams}
    \begin{tabular}{l|c|ccc}
        \toprule[1pt]
        \multicolumn{1}{c|}{\multirow{2}{*}{Method}} & \multirow{2}{*}{Params(M)} & \multicolumn{3}{c}{FLOPS (G)} \\
        \multicolumn{1}{c|}{} & & \multicolumn{1}{l}{Min.} & \multicolumn{1}{l}{Avg.} & \multicolumn{1}{l}{Max.} \\ \midrule
        Total3D & 26.556 & \textbf{16.785} & \textbf{19.199} & \textbf{31.451} \\ 
        Im3D & 26.556 & 34.004 & 47.997 & 58.023 \\ \midrule
        \textbf{Ours} & \textbf{25.599} & 22.133 & 30.201 & 44.987 \\ \bottomrule[1pt]
    \end{tabular}
\end{table}

The object detector of \emph{Implicit3D} and \emph{Total3D} are the same, which is comprised of ResNet and MLP layers. Thus their network parameters are entirely identical. While in our implementation, we use relatively fewer MLPs with fewer parameters. Regarding floating point operations, different scenes can have different FLOPS because of the different number of detected objects. We report these three methods' minimum, average and maximum FLOPS by predicting across the entire dataset. Although \emph{Implicit3D} has reported higher accuracy than ours, the proposed algorithm in this paper consumes significantly less average FLOPS, indicating that our sparse network captures more useful information and improves computational efficiency through graph pruning. In fact, our method has great potential when reasoning over complicated, cluttered indoor scenes with large scene graphs.

\section{Conclusion}
We develop a sparse graph-based approach to perform relative transformation reasoning for single image 3D object detection. The dynamic graph pruning module leverages geometric features and semantic embedding to prune the fully connected graph to a sparse graph. And the edge message update scheme incorporates both independent prediction and homogeneous transformation inference to boost the overall detection results. Extensive experiments in the paper show that our method improves the detection performance on both 3D object detection and semantic reconstruction tasks over other implicit or dense modelling methods. One limitation of our approach is that we only exploit pairwise transformation among the objects. Pursuing a more structured graph incorporating higher-order relations remains a valuable topic to be studied. It would also be promising to adapt the sparse graph model to mesh generation and layout estimation to form an end-to-end scene understanding pipeline.

\ifCLASSOPTIONcaptionsoff
  \newpage
\fi



%

\bibliographystyle{ieeetr}
\bibliography{bare_jrnl}

\begin{thebibliography}{10}

\bibitem{lian2022exploring}
Q.~Lian, B.~Ye, R.~Xu, W.~Yao, and T.~Zhang, ``Exploring geometric consistency
  for monocular 3d object detection,'' in {\em Proceedings of the IEEE/CVF
  Conference on Computer Vision and Pattern Recognition (CVPR)},
  pp.~1685--1694, June 2022.

\bibitem{liu2022learning}
X.~Liu, N.~Xue, and T.~Wu, ``Learning auxiliary monocular contexts helps
  monocular 3d object detection,'' in {\em Proceedings of the AAAI Conference
  on Artificial Intelligence}, vol.~36, pp.~1810--1818, 2022.

\bibitem{zhang2022dimension}
Y.~Zhang, W.~Zheng, Z.~Zhu, G.~Huang, D.~Du, J.~Zhou, and J.~Lu, ``Dimension
  embeddings for monocular 3d object detection,'' in {\em Proceedings of the
  IEEE/CVF Conference on Computer Vision and Pattern Recognition (CVPR)},
  pp.~1589--1598, June 2022.

\bibitem{genova2020local}
K.~Genova, F.~Cole, A.~Sud, A.~Sarna, and T.~Funkhouser, ``Local deep implicit
  functions for 3d shape,'' in {\em Proceedings of the IEEE/CVF Conference on
  Computer Vision and Pattern Recognition}, pp.~4857--4866, 2020.

\bibitem{nie2020total3dunderstanding}
Y.~Nie, X.~Han, S.~Guo, Y.~Zheng, J.~Chang, and J.~J. Zhang,
  ``Total3dunderstanding: Joint layout, object pose and mesh reconstruction for
  indoor scenes from a single image,'' in {\em Proceedings of the IEEE/CVF
  Conference on Computer Vision and Pattern Recognition}, pp.~55--64, 2020.

\bibitem{mildenhall2020nerf}
B.~Mildenhall, P.~P. Srinivasan, M.~Tancik, J.~T. Barron, R.~Ramamoorthi, and
  R.~Ng, ``Nerf: Representing scenes as neural radiance fields for view
  synthesis,'' in {\em European conference on computer vision}, pp.~405--421,
  Springer, 2020.

\bibitem{huang2018cooperative}
S.~Huang, S.~Qi, Y.~Xiao, Y.~Zhu, Y.~N. Wu, and S.-C. Zhu, ``Cooperative
  holistic scene understanding: Unifying 3d object, layout, and camera pose
  estimation,'' in {\em Advances in Neural Information Processing Systems},
  vol.~31, Curran Associates, Inc., 2018.

\bibitem{choi2013understanding}
W.~Choi, Y.-W. Chao, C.~Pantofaru, and S.~Savarese, ``Understanding indoor
  scenes using 3d geometric phrases,'' in {\em Proceedings of the IEEE
  Conference on Computer Vision and Pattern Recognition}, pp.~33--40, 2013.

\bibitem{huang2018holistic}
S.~Huang, S.~Qi, Y.~Zhu, Y.~Xiao, Y.~Xu, and S.-C. Zhu, ``Holistic 3d scene
  parsing and reconstruction from a single rgb image,'' in {\em Proceedings of
  the European Conference on Computer Vision (ECCV)}, pp.~187--203, 2018.

\bibitem{zhang2021holistic}
C.~Zhang, Z.~Cui, Y.~Zhang, B.~Zeng, M.~Pollefeys, and S.~Liu, ``Holistic 3d
  scene understanding from a single image with implicit representation,'' in
  {\em Proceedings of the IEEE/CVF Conference on Computer Vision and Pattern
  Recognition}, pp.~8833--8842, 2021.

\bibitem{chen2019holistic++}
Y.~Chen, S.~Huang, T.~Yuan, S.~Qi, Y.~Zhu, and S.-C. Zhu, ``Holistic++ scene
  understanding: Single-view 3d holistic scene parsing and human pose
  estimation with human-object interaction and physical commonsense,'' in {\em
  Proceedings of the IEEE/CVF International Conference on Computer Vision},
  pp.~8648--8657, 2019.

\bibitem{avetisyan2020scenecad}
A.~Avetisyan, T.~Khanova, C.~Choy, D.~Dash, A.~Dai, and M.~Nie{\ss}ner,
  ``Scenecad: Predicting object alignments and layouts in rgb-d scans,'' in
  {\em European Conference on Computer Vision}, pp.~596--612, Springer, 2020.

\bibitem{shi2020point}
W.~Shi and R.~Rajkumar, ``Point-gnn: Graph neural network for 3d object
  detection in a point cloud,'' in {\em Proceedings of the IEEE/CVF conference
  on computer vision and pattern recognition}, pp.~1711--1719, 2020.

\bibitem{kulkarni20193d}
N.~Kulkarni, I.~Misra, S.~Tulsiani, and A.~Gupta, ``3d-relnet: Joint object and
  relational network for 3d prediction,'' in {\em Proceedings of the IEEE/CVF
  International Conference on Computer Vision}, pp.~2212--2221, 2019.

\bibitem{zhou2019scenegraphnet}
Y.~Zhou, Z.~While, and E.~Kalogerakis, ``Scenegraphnet: Neural message passing
  for 3d indoor scene augmentation,'' in {\em Proceedings of the IEEE/CVF
  International Conference on Computer Vision}, pp.~7384--7392, 2019.

\bibitem{yang2018graph}
J.~Yang, J.~Lu, S.~Lee, D.~Batra, and D.~Parikh, ``Graph r-cnn for scene graph
  generation,'' in {\em Proceedings of the European conference on computer
  vision (ECCV)}, pp.~670--685, 2018.

\bibitem{Patil2022CVPR}
V.~Patil, C.~Sakaridis, A.~Liniger, and L.~Van~Gool, ``P3depth: Monocular depth
  estimation with a piecewise planarity prior,'' in {\em Proceedings of the
  IEEE/CVF Conference on Computer Vision and Pattern Recognition (CVPR)},
  pp.~1610--1621, June 2022.

\bibitem{qi2017pointnet}
C.~R. Qi, H.~Su, K.~Mo, and L.~J. Guibas, ``Pointnet: Deep learning on point
  sets for 3d classification and segmentation,'' in {\em Proceedings of the
  IEEE conference on computer vision and pattern recognition}, pp.~652--660,
  2017.

\bibitem{shi2020pv}
S.~Shi, C.~Guo, L.~Jiang, Z.~Wang, J.~Shi, X.~Wang, and H.~Li, ``Pv-rcnn:
  Point-voxel feature set abstraction for 3d object detection,'' in {\em
  Proceedings of the IEEE/CVF Conference on Computer Vision and Pattern
  Recognition}, pp.~10529--10538, 2020.

\bibitem{shi2019pointrcnn}
S.~Shi, X.~Wang, and H.~Li, ``Pointrcnn: 3d object proposal generation and
  detection from point cloud,'' in {\em Proceedings of the IEEE/CVF Conference
  on Computer Vision and Pattern Recognition}, pp.~770--779, 2019.

\bibitem{tmm22_detection}
S.~Jiayao, S.~Zhou, Y.~Cui, and Z.~Fang, ``Real-time 3d single object tracking
  with transformer,'' {\em IEEE Transactions on Multimedia}, pp.~1--1, 2022.

\bibitem{chen2017multi}
X.~Chen, H.~Ma, J.~Wan, B.~Li, and T.~Xia, ``Multi-view 3d object detection
  network for autonomous driving,'' in {\em Proceedings of the IEEE conference
  on Computer Vision and Pattern Recognition}, pp.~1907--1915, 2017.

\bibitem{qi2018frustum}
C.~R. Qi, W.~Liu, C.~Wu, H.~Su, and L.~J. Guibas, ``Frustum pointnets for 3d
  object detection from rgb-d data,'' in {\em Proceedings of the IEEE
  conference on computer vision and pattern recognition}, pp.~918--927, 2018.

\bibitem{tmm20_layout}
C.~Yan, B.~Shao, H.~Zhao, R.~Ning, Y.~Zhang, and F.~Xu, ``3d room layout
  estimation from a single rgb image,'' {\em IEEE Transactions on Multimedia},
  vol.~22, no.~11, pp.~3014--3024, 2020.

\bibitem{tmm21_layout}
W.~Zhang, Q.~Zhang, W.~Zhang, J.~Gu, and Y.~Li, ``From edge to keypoint: An
  end-to-end framework for indoor layout estimation,'' {\em IEEE Transactions
  on Multimedia}, vol.~23, pp.~4483--4490, 2021.

\bibitem{shin20193d}
D.~Shin, Z.~Ren, E.~B. Sudderth, and C.~C. Fowlkes, ``3d scene reconstruction
  with multi-layer depth and epipolar transformers,'' in {\em Proceedings of
  the IEEE/CVF International Conference on Computer Vision}, pp.~2172--2182,
  2019.

\bibitem{wu2020pq}
R.~Wu, Y.~Zhuang, K.~Xu, H.~Zhang, and B.~Chen, ``Pq-net: A generative part
  seq2seq network for 3d shapes,'' in {\em Proceedings of the IEEE/CVF
  Conference on Computer Vision and Pattern Recognition}, pp.~829--838, 2020.

\bibitem{xu2019disn}
Q.~Xu, W.~Wang, D.~Ceylan, R.~Mech, and U.~Neumann, ``Disn: Deep implicit
  surface network for high-quality single-view 3d reconstruction,'' {\em
  Advances in Neural Information Processing Systems}, vol.~32, 2019.

\bibitem{izadinia2017im2cad}
H.~Izadinia, Q.~Shan, and S.~M. Seitz, ``Im2cad,'' in {\em Proceedings of the
  IEEE Conference on Computer Vision and Pattern Recognition}, pp.~5134--5143,
  2017.

\bibitem{wang2018pixel2mesh}
N.~Wang, Y.~Zhang, Z.~Li, Y.~Fu, W.~Liu, and Y.-G. Jiang, ``Pixel2mesh:
  Generating 3d mesh models from single rgb images,'' in {\em Proceedings of
  the European Conference on Computer Vision (ECCV)}, pp.~52--67, 2018.

\bibitem{gkioxari2019mesh}
G.~Gkioxari, J.~Malik, and J.~Johnson, ``Mesh r-cnn,'' in {\em Proceedings of
  the IEEE/CVF International Conference on Computer Vision}, pp.~9785--9795,
  2019.

\bibitem{pan2019deep}
J.~Pan, X.~Han, W.~Chen, J.~Tang, and K.~Jia, ``Deep mesh reconstruction from
  single rgb images via topology modification networks,'' in {\em Proceedings
  of the IEEE/CVF International Conference on Computer Vision}, pp.~9964--9973,
  2019.

\bibitem{Park2019CVPR}
J.~J. Park, P.~Florence, J.~Straub, R.~Newcombe, and S.~Lovegrove, ``Deepsdf:
  Learning continuous signed distance functions for shape representation,'' in
  {\em Proceedings of the IEEE/CVF Conference on Computer Vision and Pattern
  Recognition (CVPR)}, June 2019.

\bibitem{Li2022CVPR}
T.~Li, X.~Wen, Y.-S. Liu, H.~Su, and Z.~Han, ``Learning deep implicit functions
  for 3d shapes with dynamic code clouds,'' in {\em Proceedings of the IEEE/CVF
  Conference on Computer Vision and Pattern Recognition (CVPR)},
  pp.~12840--12850, June 2022.

\bibitem{johnson2015image}
J.~Johnson, R.~Krishna, M.~Stark, L.-J. Li, D.~Shamma, M.~Bernstein, and
  L.~Fei-Fei, ``Image retrieval using scene graphs,'' in {\em Proceedings of
  the IEEE conference on computer vision and pattern recognition},
  pp.~3668--3678, 2015.

\bibitem{liu2016ssd}
W.~Liu, D.~Anguelov, D.~Erhan, C.~Szegedy, S.~Reed, C.-Y. Fu, and A.~C. Berg,
  ``Ssd: Single shot multibox detector,'' in {\em European conference on
  computer vision}, pp.~21--37, Springer, 2016.

\bibitem{redmon2016you}
J.~Redmon, S.~Divvala, R.~Girshick, and A.~Farhadi, ``You only look once:
  Unified, real-time object detection,'' in {\em Proceedings of the IEEE
  conference on computer vision and pattern recognition}, pp.~779--788, 2016.

\bibitem{he2017mask}
K.~He, G.~Gkioxari, P.~Dollar, and R.~Girshick, ``Mask r-cnn,'' in {\em
  Proceedings of the IEEE international conference on computer vision},
  pp.~2961--2969, 2017.

\bibitem{wang2019graph}
L.~Wang, Y.~Huang, Y.~Hou, S.~Zhang, and J.~Shan, ``Graph attention convolution
  for point cloud semantic segmentation,'' in {\em Proceedings of the IEEE/CVF
  conference on computer vision and pattern recognition}, pp.~10296--10305,
  2019.

\bibitem{wald2020learning}
J.~Wald, H.~Dhamo, N.~Navab, and F.~Tombari, ``Learning 3d semantic scene
  graphs from 3d indoor reconstructions,'' in {\em Proceedings of the IEEE/CVF
  Conference on Computer Vision and Pattern Recognition}, pp.~3961--3970, 2020.

\bibitem{battaglia2016interaction}
P.~Battaglia, R.~Pascanu, M.~Lai, D.~Jimenez~Rezende, and K.~Kavukcuoglu,
  ``Interaction networks for learning about objects, relations and physics,''
  in {\em Advances in Neural Information Processing Systems}, vol.~29, Curran
  Associates, Inc., 2016.

\bibitem{scarselli2008graph}
F.~Scarselli, M.~Gori, A.~C. Tsoi, M.~Hagenbuchner, and G.~Monfardini, ``The
  graph neural network model,'' {\em IEEE transactions on neural networks},
  vol.~20, no.~1, pp.~61--80, 2008.

\bibitem{zhang2020deep}
Z.~Zhang, P.~Cui, and W.~Zhu, ``Deep learning on graphs: A survey,'' {\em IEEE
  Transactions on Knowledge and Data Engineering}, 2020.

\bibitem{gilmer2017neural}
J.~Gilmer, S.~S. Schoenholz, P.~F. Riley, O.~Vinyals, and G.~E. Dahl, ``Neural
  message passing for quantum chemistry,'' in {\em International Conference on
  Machine Learning}, pp.~1263--1272, PMLR, 2017.

\bibitem{kipf2016semi}
T.~N. Kipf and M.~Welling, ``Semi-supervised classification with graph
  convolutional networks,'' {\em arXiv preprint arXiv:1609.02907}, 2016.

\bibitem{vaswani2017attention}
A.~Vaswani, N.~Shazeer, N.~Parmar, J.~Uszkoreit, L.~Jones, A.~N. Gomez,
  L.~Kaiser, and I.~Polosukhin, ``Attention is all you need,'' {\em arXiv
  preprint arXiv:1706.03762}, 2017.

\bibitem{hu2018relation}
H.~Hu, J.~Gu, Z.~Zhang, J.~Dai, and Y.~Wei, ``Relation networks for object
  detection,'' in {\em Proceedings of the IEEE Conference on Computer Vision
  and Pattern Recognition}, pp.~3588--3597, 2018.

\bibitem{devlin-etal-2019-bert}
J.~Devlin, M.-W. Chang, K.~Lee, and K.~Toutanova, ``{BERT}: Pre-training of
  deep bidirectional transformers for language understanding,'' in {\em
  Proceedings of the 2019 Conference of the North {A}merican Chapter of the
  Association for Computational Linguistics: Human Language Technologies,
  Volume 1 (Long and Short Papers)}, pp.~4171--4186, 2019.

\bibitem{campa2009acc}
R.~Campa and H.~de~la Torre, ``Pose control of robot manipulators using
  different orientation representations: A comparative review,'' in {\em 2009
  American Control Conference}, pp.~2855--2860, 2009.

\bibitem{ren2015faster}
S.~Ren, K.~He, R.~Girshick, and J.~Sun, ``Faster r-cnn: Towards real-time
  object detection with region proposal networks,'' in {\em Advances in Neural
  Information Processing Systems}, vol.~28, Curran Associates, Inc., 2015.

\bibitem{song2015sun}
S.~Song, S.~P. Lichtenberg, and J.~Xiao, ``Sun rgb-d: A rgb-d scene
  understanding benchmark suite,'' in {\em Proceedings of the IEEE conference
  on computer vision and pattern recognition}, pp.~567--576, 2015.

\bibitem{silberman2012indoor}
N.~Silberman, D.~Hoiem, P.~Kohli, and R.~Fergus, ``Indoor segmentation and
  support inference from rgbd images,'' in {\em European conference on computer
  vision}, pp.~746--760, Springer, 2012.

\bibitem{sun2018pix3d}
X.~Sun, J.~Wu, X.~Zhang, Z.~Zhang, C.~Zhang, T.~Xue, J.~B. Tenenbaum, and W.~T.
  Freeman, ``Pix3d: Dataset and methods for single-image 3d shape modeling,''
  in {\em Proceedings of the IEEE Conference on Computer Vision and Pattern
  Recognition}, pp.~2974--2983, 2018.

\bibitem{tulsiani2018factoring}
S.~Tulsiani, S.~Gupta, D.~F. Fouhey, A.~A. Efros, and J.~Malik, ``Factoring
  shape, pose, and layout from the 2d image of a 3d scene,'' in {\em
  Proceedings of the IEEE Conference on Computer Vision and Pattern
  Recognition}, pp.~302--310, 2018.

\bibitem{groueix2018papier}
T.~Groueix, M.~Fisher, V.~G. Kim, B.~C. Russell, and M.~Aubry, ``A
  papier-m{\^a}ch{\'e} approach to learning 3d surface generation,'' in {\em
  Proceedings of the IEEE conference on computer vision and pattern
  recognition}, pp.~216--224, 2018.

\end{thebibliography}


\begin{thebibliography}{1}
\bibitem{IEEEhowto:kopka}
H.~Kopka and P.~W. Daly, \emph{A Guide to \LaTeX}, 3rd~ed.\hskip 1em plus
  0.5em minus 0.4em\relax Harlow, England: Addison-Wesley, 1999.
\end{thebibliography}

%

\end{document}